\documentclass[lettersize,journal]{IEEEtran}
\usepackage{amsmath,amsfonts}
\usepackage{algorithmic}
\usepackage{algorithm}
\usepackage{array}
\usepackage{textcomp}
\usepackage{stfloats}
\usepackage{url}
\usepackage{color}
\usepackage{subfigure}
\usepackage{makecell}
\usepackage{multirow}
\usepackage{verbatim}
\usepackage{graphicx}
\usepackage{mathrsfs}
\usepackage{bbding}
\usepackage{cite}
\usepackage{enumitem}
\usepackage{amsmath,amssymb,amsthm}
\usepackage{gensymb}
\usepackage{dsfont}
\usepackage{threeparttable}
\usepackage{booktabs}
\usepackage{colortbl}
\usepackage[table,xcdraw]{xcolor}
\usepackage{url}
\usepackage[colorlinks,
            linkcolor=black,
            anchorcolor=black,
            citecolor=black]{hyperref}

\begin{document}

\title{
Seeing to Act, Prompting to Specify: A Bayesian Factorization of Vision Language Action Policy
}

\author{Kechun Xu\IEEEauthorrefmark{4}\IEEEauthorrefmark{5}, Zhenjie Zhu\IEEEauthorrefmark{4}, Anzhe Chen\IEEEauthorrefmark{4}, Shuqi Zhao\IEEEauthorrefmark{5}, Qing Huang\IEEEauthorrefmark{4}, Yifei Yang\IEEEauthorrefmark{4}, \\ Haojian Lu\IEEEauthorrefmark{4}, Rong Xiong\IEEEauthorrefmark{4}, Masayoshi Tomizuka\IEEEauthorrefmark{5}\IEEEauthorrefmark{3}, Yue Wang\IEEEauthorrefmark{4}\IEEEauthorrefmark{3}
\thanks{\IEEEauthorrefmark{4} Zhejiang University, \IEEEauthorrefmark{5} UC Berkeley. \IEEEauthorrefmark{3}Corresponding authors: Masayoshi Tomizuka, {\tt\small tomizuka@me.berkeley.edu} and Yue Wang, {\tt\small wangyue@iipc.zju.edu.cn}. }
}

\maketitle

\begin{abstract}
The pursuit of out-of-distribution generalization in Vision–Language–Action~(VLA) models is often hindered by catastrophic forgetting of the Vision-Language Model~(VLM) backbone during fine-tuning. While co-training with external reasoning data helps, it requires experienced tuning and data-related overhead. Beyond such external dependencies, we identify an intrinsic cause within VLA datasets: \textit{modality imbalance}, where language diversity is much lower than visual and action diversity. This imbalance biases the model toward visual shortcuts and language forgetting. To address this, we introduce BayesVLA, a Bayesian factorization that decomposes the policy into a visual–action prior, supporting \textit{seeing-to-act}, and a language-conditioned likelihood, enabling \textit{prompt-to-specify}. This inherently preserves generalization and promotes instruction following. We further incorporate pre- and post-contact phases to better leverage pre-trained foundation models. Information-theoretic analysis formally validates our effectiveness in mitigating shortcut learning. Extensive experiments show superior generalization to unseen instructions, objects, and environments compared to existing methods. Project page is available at: \href{https://xukechun.github.io/papers/BayesVLA}{https://xukechun.github.io/papers/BayesVLA}.
\end{abstract}

\section{Introduction}

\IEEEPARstart{E}{nabling} robots to perform diverse manipulation tasks in various environments remains a fundamental challenge in robotics. The emergence of learning-based Vision–Language–Action (VLA) models offers a promising direction by integrating visual perception with natural language instructions to generate robot actions~\cite{brohan2022rt1,team2024octo,zhong2025surveyvla}. For these models, a critical requirement is out-of-distribution~(OOD) generalization, which is the ability to handle unseen task-level prompt variations.

To build such generalizable policies, a common approach~\cite{zitkovich2023rt2,kim2024openvla,kim2025oft,bjorck2025gr00tn1,black2024pi0} leverages pre-trained Vision–Language Models (VLMs)~\cite{radford2021learning,beyer2024paligemma,wang2024qwen2vl} as backbones, which are first pre-trained on diverse robot datasets~\cite{openxembodiment,wu2024robomind,bu2025agibotworld} and then fine-tuned on task-specific data. However, a primary obstacle in this paradigm is \emph{catastrophic forgetting}~\cite{liu2023libero,gao2025taxonomy}, {\it i.e.}, fine-tuning overwrites the pre-trained representations that are essential for generalization. As a result, the policy performs well on seen instructions but loses the ability to follow unseen ones. Co-training VLAs with both reasoning data and robot data has proven effective in mitigating catastrophic forgetting and preserving generalization~\cite{wen2025dexvla,wen2025diffusionvla,yang2025instructvla}. However, this approach relies on heuristic tuning of the data ratio between reasoning and action domains, as well as the careful selection of fine-tuning techniques. In addition, for end-user deployment, a prohibitive cost is needed for collecting, storing, and training the reasoning data. Therefore, we raise the question: \textit{Can we develop a method that improves instruction-following generalization without relying on external reasoning data, while being guided by a more principled framework?}

We argue that the success of co-training, while valid, has obscured a deeper issue: the inherent modality imbalance in VLA datasets. This structural problem is characterized by a stark disparity in diversity, where \textit{language instructions are significantly less diverse than the corresponding visual observations and actions}. We posit that this imbalance is the underlying reason why heuristic tuning of the data ratio between reasoning and action domains is critical during co-training. Without reasoning data, this imbalance further exacerbates catastrophic forgetting, as the model is incentivized to discard nuanced language understanding in favor of shortcut visual features, leading to severe degradation in language generalization.

\begin{figure}[t]
  \centering
  \includegraphics[width=\linewidth]{images/teaser.pdf}
  \caption{
  Compared to typical VLA policies, BayesVLA decomposes the policy into a vision-action prior and a language-conditioned likelihood. The vision–action prior leverages visual information for action generation~({\it seeing to act}), while the language-conditioned likelihood aligns these action priors with the language instruction~({\it prompting to specify}). 
  }
  \label{fig:teaser}
\end{figure}

In response to this structural disparity, we propose \emph{Bayesian factorization} as a principled alternative to heuristic tuning of co-training. We formally decompose the VLA policy into a vision–action prior and a language-conditioned likelihood. This probabilistic formulation turns the modality imbalance into the architecture: a robust vision–action pathway leverages abundant visuomotor data to learn manipulation primitives, while a separate language pathway is dedicated to instruction specification. By structurally addressing the data imbalance during fine-tuning, our approach mitigates forgetting of language understanding and ensures robust task-level generalization performance without reliance on external reasoning data.

In this paper, we propose \textbf{BayesVLA}, which models the VLA policy using a prior and a likelihood. As shown in Fig.~\ref{fig:teaser}, this factorization leads to a natural two-stage training procedure: the language-agnostic prior first learns foundational visuomotor control, while the language-conditioned likelihood then prompts the model to specify actions conditioned on instructions. This design encourages the model to leverage language cues without distorting visual representations, inherently mitigating language forgetting. Furthermore, inspired by the decomposition of general object manipulation~\cite{koval2016prepostcontact}, we structure the policy into pre-contact and post-contact phases. For the pre-contact phase, we employ the pre-trained foundation models of specific primitives for candidate action generation~\cite{fang2023anygrasp,zhao2025anyplace}. For the post-contact phase, we train a diffusion-based action generator to produce multimodal trajectories under vision observations. Then, VLM features are injected in the second training stage for both phases. Leveraging these pre-trained foundation models, we relax the need for unaffordable large-scale pre-training. We further provide an information-theoretic analysis that directly links data bias to poor instruction following and validates the effectiveness of our Bayesian factorization. The generalization capability of our policy is evaluated on public benchmarks and self-built simulation environments, spanning challenges such as unseen objects and scenes. Extensive real-world experiments confirm that BayesVLA outperforms existing methods in language-conditioned generalization. To summarize, the main contributions are:
\begin{itemize}
    \item Bayesian factorization framework for structurally handling modality imbalance in VLA training.
    \item BayesVLA consisting of specialized pre-contact and post-contact policies. 
    \item Novel techniques for modeling the likelihood and aligning the prior.
    \item Information-theoretical analysis validating the effectiveness of factorization.
    \item Comprehensive evaluation demonstrating superior generalization in both simulation and real-world scenarios.
\end{itemize}

\section{Related Works}

\subsection{Vision and Language Foundation Models for Manipulation}

Foundation models in computer vision and natural language processing have shown remarkable capabilities~\cite{radford2021learning,hurst2024gpt4o,wang2024qwen2vl,beyer2024paligemma,abdolmaleki2025geminirobotics1_5,bai2025qwen2_5}, motivating their application to generalizable robotic manipulation in open-world settings~\cite{firoozi2023foundation}. A common strategy is to directly ground these models in robotic scenarios. Several approaches~\cite{Ahn2022DoAI,huang2023instruct2act,xu2024gsp,zhu2025grounding} leverage vision foundation models for object grounding from flexible language instructions. Among them, some focus on object-centric representations~\cite{xu2023vilg,zhu2023viola}, while others construct 3D scene representations that jointly encode semantic and geometric information~\cite{wang2024d,rashid2023lerftogo,xu2025a2}.
In parallel, a series of works~\cite{liang2023code,huang2023voxposer,duan2024manipulate,lou2025explorevlm} exploits the reasoning abilities of Large Language Models~(LLMs) to develop high-level task planners. However, the performance of such systems often suffers from cascaded errors across perception, reasoning, and control modules. Another line of research~\cite{jiang2023vima,ke20243d,goyal2024rvt2,fang2025sam2act} integrates features from vision foundation models into end-to-end policy learning, achieving promising results but requiring extensive demonstrations and long training horizons to converge. Moreover, these methods typically struggle to generalize when faced with objects or environments that differ significantly from their training distribution.

\subsection{Action Foundation Models for Manipulation}

Object manipulation is commonly divided into pre-contact and post-contact phases~\cite{koval2016prepostcontact}. Following this perspective, recent efforts have explored action foundation models trained on large-scale robotic datasets to generate actions across diverse scenarios. Most existing models specialize in specific action primitives~\cite{fang2023anygrasp,vuong2024graspanything,zhao2025anyplace,xu2022eom,NoeMatrixAnySkill}. For example, AnyGrasp~\cite{fang2023anygrasp} develops a grasp foundation model capable of predicting diverse grasp poses, while AnyPlace~\cite{zhao2025anyplace} develops a unified placement policy. These models typically output multiple feasible contact poses conditioned on visual observation, whose annotations are easier to obtain and scale. Consequently, they exhibit strong robustness and generalization across varied scenarios. Building on these primitive-level models, recent work aims to achieve multi-task manipulation through unified policies that generalize across tasks and embodiments, such as ACT and DP~\cite{jang2022bc-z,zhao2023act,chi2023dp,ze2024dp3,zhu2025scaledp}. These approaches predict dense action trajectories and support a wide range of manipulation behaviors. This progression has catalyzed a growing interest in large-scale Vision-Language-Action~(VLA) models~\cite{openxembodiment,kim2024openvla,black2024pi0}, which integrate visual grounding, language understanding, and action generation within a single framework. However, research on action foundation models and VLA models remains largely disconnected. In particular, how to effectively leverage the strong priors provided by action foundation models within VLA frameworks is still underexplored. Existing attempts~\cite{rashid2023lerftogo,yenamandra2023homerobot} typically invoke a pre-trained grasp model only after object grounding, without deeper integration. More recent studies investigate steering pre-trained policies using value functions~\cite{nakamoto2024steering,wu2025foresight,du2025dynaguide}. In this paper, we bridge these developments by introducing a unified framework that incorporates priors from vision, language, and action foundation models, enabling improved generalization in challenging manipulation scenarios. 

\subsection{Vision-Language-Action Models}

To build general-purpose robot policies capable of performing diverse tasks and generalizing to new settings, recent studies expand training data from single-task to multi-scene, multi-task datasets~\cite{bharadhwaj2024roboagent,openxembodiment,khazatsky2024droid,fang2024rh20t,wu2024robomind,bu2025agibotworld,chen2025robotwin2,geng2025roboverse,generalist2025gen0}. Among these efforts, Vision-Language-Action~(VLA) models have emerged as a central paradigm, integrating perception, language understanding, and control into a unified framework~\cite{zhong2025surveyvla,liu2025robovlms,cui2025openhelix}. Early representative works such as RT-1 and Octo~\cite{brohan2022rt1, team2024octo} train transformer-based policies from scratch using large-scale robot data. Subsequent advances~\cite{zitkovich2023rt2,rt-traj,kim2024openvla,pertsch2025fast,zhao2025cotvla,huangotter} incorporate pre-trained Vision-Language Models~(VLMs)~\cite{anschutz2023prismatic,beyer2024paligemma} and fine-tune them on robot data via action discretization, leveraging VLM priors for improved visual and task grounding. However, the inference latency of these models often limits their deployment in real-time robotic control. To overcome this issue, a line of recent work~\cite{liu2024rdt,wen2025dexvla,wen2025diffusionvla,hou2025dita,kim2025oft,rdt2,yang2025instructvla,liu2025hybridvla} adopts diffusion-based models for continuous multimodal action representation, enabling smoother and more expressive policy outputs. In parallel, dual-system architectures are proposed to decouple high-level reasoning from low-level control, where a slow reasoning module~(typically VLM-based backbone) operates alongside a fast action expert based on diffusion or transformer backbones~\cite{black2024pi0,bjorck2025gr00tn1,bu2024robodual,shukor2025smolvla,deng2025graspvla,jiang2025rynnvla,barreiros2025lbm,bu2025univla,jiang2025galaxea,cen2025worldvla,zhang2025dreamvla}. These dual systems perform both task grounding and action generation and have become a dominant trend in VLA research.

Despite these advances, most existing policies still exhibit limited generalization, particularly when faced with novel objects, unseen language instructions, or new task compositions~\cite{liu2023libero,gao2025taxonomy,zhou2025liberopro}.
Several studies attempt to preserve the generalization of pre-trained VLM by freezing it during policy fine-tuning~\cite{bjorck2025gr00tn15,driess2025knowledge}, or by co-training VLM with robot data~\cite{black2025pi_ohfive,cheang2025gr3}, yielding improved instruction following generalization. Nevertheless, the performance of these approaches relies on heuristic tuning of updated parameters or data ratio between reasoning and action data. In this work, we take a structural perspective on policy learning by decomposing action generation and language alignment under a Bayesian formulation, enabling better generalization across unseen objects and tasks.

\section{Bayesian Modeling}

\subsection{Preliminary}

{

{\bf Vision-Language-Action Formulation.} We denote the Vision-Language-Action (VLA) policy as $\pi(\mathbf{a}|\mathbf{v},\ell)$, where $\mathbf{a}$ is the robot action, $\mathbf{v}$ is the current visual observations, and $\ell$ is the language instruction. We assume a dataset $D_{\mathrm{VLA}}$ that contains triplets $(\mathbf{a}, \mathbf{v}, \ell)$ for training the VLA policy. The corresponding training objective is
\begin{equation}
    \min -\mathbb{E}_{D_\mathrm{VLA}} [\log\pi(\mathbf{a}|\mathbf{v},\ell)]
\end{equation}
where $\mathbb{E}$ is the expectation over the dataset $D_\mathrm{VLA}$. 

Typically, VLMs are employed as the backbone of $\pi(\mathbf{a}|\mathbf{v},\ell)$~\cite{zitkovich2023rt2,kim2024openvla}, since their vision-language knowledge is expected to provide better generalization. Besides, as VLMs produce discrete token outputs, recent works introduce an action expert that takes the VLM backbone tokens as input to generate continuous actions~\cite{black2024pi0,bjorck2025gr00tn1}. There is no pre-trained action expert, so it is trained from scratch jointly with the pre-trained VLM backbone.

{\bf Catastrophic Forgetting and Co-training.}  Training the action expert typically requires many optimization iterations, which can lead to catastrophic forgetting in the VLM backbone and degrade its generalization ability.  A straightforward solution is to jointly train on VLM reasoning data $D_{\mathrm{VLM}}$ along with VLA data, following the common co-training paradigm. In this way, the reasoning loss helps mitigate catastrophic forgetting:
\begin{equation}
    \min -\mathbb{E}_{D_\mathrm{VLA}} [\log\pi(\mathbf{a}|\mathbf{v},\ell)] - \mathbb{E}_{D_\mathrm{VLM}} [\log \mathcal{V}(\ell'|\mathbf{v},\ell)]
\end{equation}
where $\mathcal{V}$ denotes the VLM backbone, and $\ell'$ is the next token prediction target for vision-language reasoning. Such co-training is proven effective in alleviating forgetting. However, when end-users aim to fine-tune a VLA model for a specific task, they must collect, store, and train on $D_{\mathrm{VLM}}$, which significantly increases deployment cost. Parameter-efficient fine-tuning~(PEFT)~\cite{hu2022lora} can reduce this overhead, but it still relies on tuning experience \textit{e.g.} layer of VLM for adaptation.

{\bf Data Imbalance.} Beyond the need for external reasoning data $D_{\mathrm{VLM}}$, we identify a more fundamental limitation within standard VLA datasets $D_\mathrm{VLA}$. We note that the triplets in these datasets are inherently imbalanced~\cite{wanna2025lang}. For each VLA demonstration, we have many vision-action pairs, say $T$ frames $\{ \mathbf{a}_{1\cdots T}, \mathbf{v}_{1\cdots T}\}$. But the language instruction for this video is unique. As a result, the $T$ triplets take the form $\{\mathbf{a}_{1\cdots T}, \mathbf{v}_{1\cdots T}, \ell\}$, where the visual and action modalities exhibit $T$-fold higher diversity compared to the language modality. This structural bias encourages the model to exploit statistical shortcuts, learning to predict actions primarily from visual cues while ignoring language instructions. In Sec.~\ref{sec:inform-theory}, we theoretically demonstrate how this imbalance systematically undermines language grounding and induces a bias toward vision-only policies.
}

\subsection{Bayesian Factorization}
\begin{figure*}[t]
  \centering
  \includegraphics[width=\textwidth]{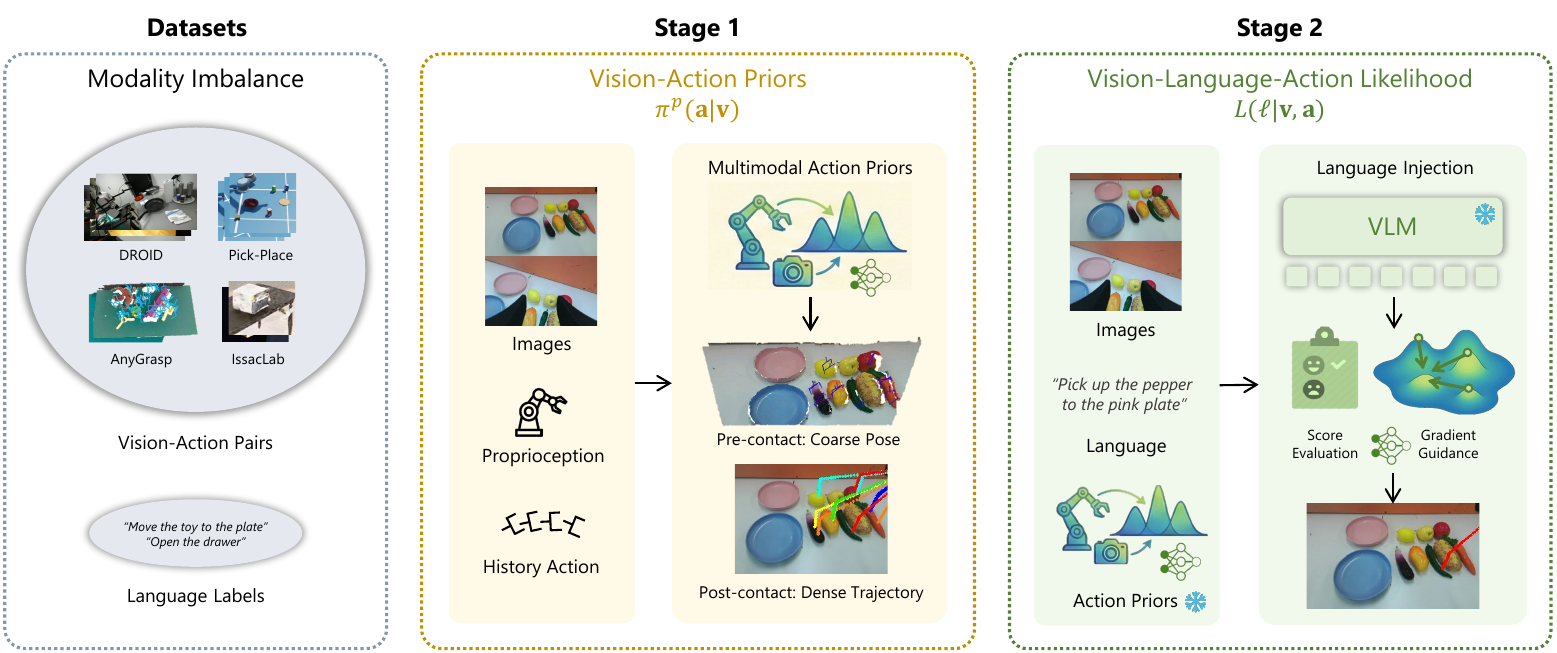}
  \caption{
  {\bf Overview}. Given VLA datasets with modality imbalance, BayesVLA models the VLA policy using a prior and a likelihood. For stage 1, we train a prior model $\pi^{p}(\mathbf{a}|\mathbf{v})$ that takes in visual input to generate multimodal action distribution. Based on the prior, for stage 2, we train the likelihood model to further align the action priors with the language instruction.}
  \label{fig:overview}
\end{figure*}
To address the dataset bias, we propose \textit{see-to-act then prompt-to-specify}, an internal policy structure instead of depending on external reasoning data, to ensure more robust language grounding. We present a Bayesian factorization of the VLA policy, which induces a principled two-stage architecture: first generating action priors from vision, then refining them via language-aligned likelihood. Formally, this factorization is expressed as: 
\begin{equation}
\pi(\mathbf{a}|\mathbf{v},\ell) \propto \pi^{p}(\mathbf{a}|\mathbf{v}) \, L(\ell|\mathbf{v},\mathbf{a})
\label{eq:bayesian}
\end{equation}
where $\pi^p(\mathbf{a}|\mathbf{v})$ denotes the vision-action prior, and $L(\ell|\mathbf{v},\mathbf{a})$ represents the language-conditioned likelihood.

{\bf VA Prior.} The prior model $\pi^p(\mathbf{a}|\mathbf{v})$ captures the action distribution conditioned on visual observations, representing multimodal action affordances such as grasp poses or trajectories. This model encodes what can be done under current observation, providing action priors for downstream specific tasks.

{\bf VLA Likelihood.} Based on the prior $\pi^p(\mathbf{a}|\mathbf{v})$, the likelihood model $L(\ell|\mathbf{v},\mathbf{a})$ prunes the action distribution under the guidance of specific task information, {\it e.g.} language instruction. This model grounds language in action, providing language-aligned behaviors for execution.

Together, the prior and likelihood constitute a VLA policy, unifying perception, language understanding, and action generation within a single framework.

\subsection{Decomposed Action Generation and Language Alignment} 
Guided by this modeling, we propose {\bf BayesVLA}, which learns two models as prior and likelihood, respectively. The prior focuses on feasible action generation, while the likelihood focuses on language grounding and alignment. This factorization naturally induces a two-stage training
procedure, as illustrated in Fig.~\ref{fig:overview}.

{\bf Action Generation by Prior.} We first train the prior model $\pi^p(\mathbf{a}|\mathbf{v})$ to generate diverse action proposals conditioned on visual input. Given the visual input, the prior model predicts a multimodal action distribution that captures feasible and meaningful interactions, without requiring task-specific instructions.
This allows the model to exploit the balanced diversity in vision-action pairs, learning general action priors such as contact poses or motion trajectories.

{\bf Language Alignment by Likelihood.} Then, with the pre-trained prior, we train the likelihood model $L(\ell|\mathbf{v},\mathbf{a})$ to align action priors with language instructions, leveraging generalizable semantic knowledge from foundation models. The likelihood model evaluates the action proposals based on the language instruction, rather than generating actions from scratch. This allows the likelihood to remain compact and data-efficient, enabling it to inherit the generalization capabilities of pre-trained foundation models without overwriting their core knowledge through large-scale fine-tuning.

{\bf Generalization by Factorization.} Compared with previous work that directly learns a single end-to-end policy $\pi(\mathbf{a}|\mathbf{v},\ell)$ coupling action generation and language understanding, our Bayesian factorization provides three key advantages for better generalization. First, it decomposes the tasks of action generation and language alignment, simplifying the training of both models. Second, it mitigates modality imbalance by first learning a prior for action generation, which utilizes balanced diversity of vision-action priors. Third,  it better preserves the semantic capacity of the pre-trained foundation models through lightweight likelihood adaptation, and improves interpretability by enabling explicit action evaluation and reweighting based on language alignment.

\subsection{Phases in Object Manipulation}
In general, object manipulation naturally involves two phases~\cite{koval2016prepostcontact}: a pre-contact phase and a post-contact phase. As shown in Fig.~\ref{fig:teaser}, during the pre-contact phase, the robot reaches and contacts the object of interest. During the post-contact phase, the robot, which contacts the object, executes diverse trajectories for general manipulation. Upon these properties, action skills can be categorized into pre-contact and post-contact ones.

{\bf Pre-contact Phase.} In the pre-contact phase, the primary objective is to predict a contact pose, then the robot can reach the contact pose without physical interaction using motion planning for collision avoidance. 

{\bf Post-contact Phase}. After contacting the object, the post-contact actions involve rich interactions with the object and environment. Therefore, these actions are supposed to be continuous trajectories to capture general object manipulation skills, {\it e.g.}, placing, opening, or pouring. 

These two phases differ in action representation granularity, data availability, and multimodality. In BayesVLA, we adopt a unified prior-likelihood formulation for both phases, while instantiating different architectures to accommodate their distinct characteristics. This enables flexible employment of our framework across diverse manipulation skills.

\section{BayesVLA Architectures}

\subsection{Pre-contact Phase}

In the pre-contact phase, action generation can be simplified to predicting a 6-DoF contact pose, {\it e.g.}, grasping or pick-like actions. Once the contact pose is determined, the robot can execute motion planning to reach the contact configuration while ensuring collision avoidance. Accordingly, the prior $\pi^{p}(\mathbf{a}|\mathbf{v})$ is instantiated as a contact pose generator conditioned on visual observations. Compared to dense trajectory prediction, this formulation substantially reduces learning complexity without sacrificing manipulation performance. In this section, we introduce the prior and likelihood formulations for pre-contact phase, along with their training procedures.

\begin{figure}[t]
  \centering
  \includegraphics[width=\linewidth]{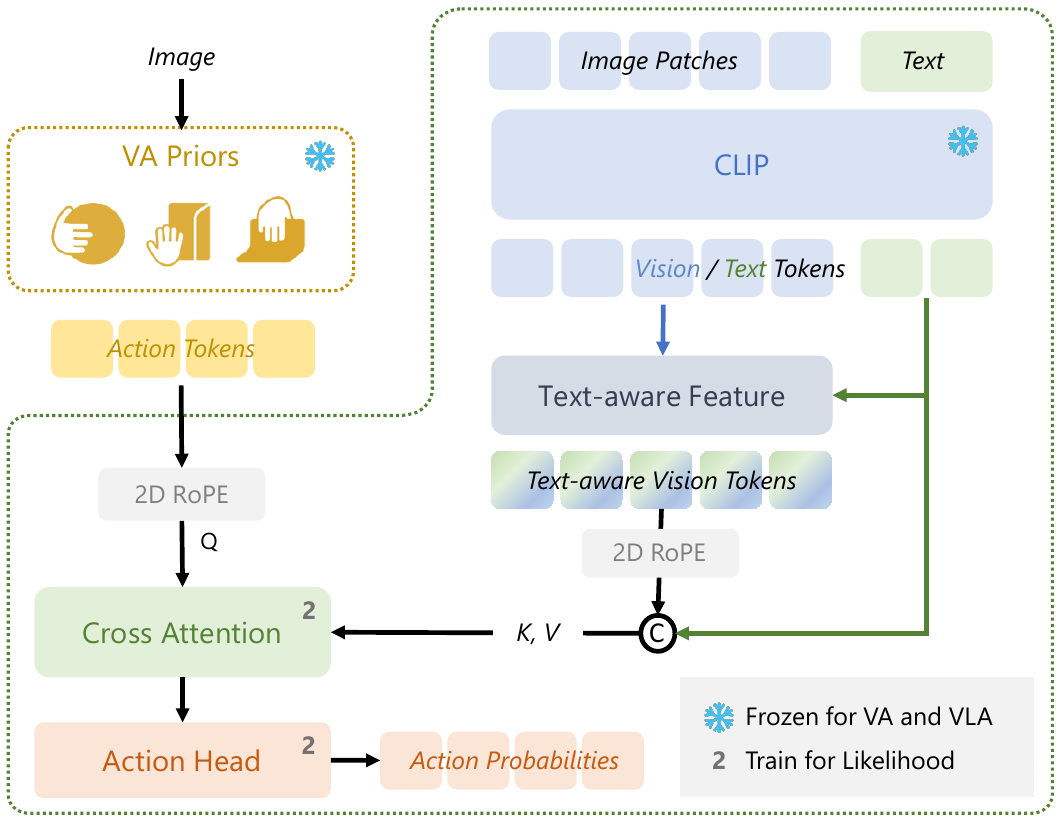}
  \caption{Architecture of pre-contact phase. We leverage pre-trained models as vision-action priors, which yield multiple contact poses as action tokens. Then, the text tokens and text-aware vision tokens are injected with cross attention layers to predict action probabilities.} 
  \label{fig:pre_arch}
\end{figure}

{\bf Pre-trained Action Foundation Models as Prior.} For pre-contact actions, there are some action foundation models~\cite{fang2023anygrasp,zhao2025anyplace,xu2022eom} trained to predict contact poses, {\it e.g.}, grasping poses. These models are trained with large-scale vision-action datasets with annotated contact poses to maximize coverage and diversity of feasible contacts. Thanks to the large-scale pre-training, they demonstrate a good capability to generate multimodal action patterns conditioned on visual observation, and generalize well to novel scenarios. Consequently, such pre-trained models can naturally be priors for contact pose generation. For instance, we adopt pre-trained grasp foundation models, {\it i.e.}, AnyGrasp~\cite{fang2023anygrasp} to predict multimodal grasp poses that are feasible for object manipulation. Consider action proposals generated by pre-trained action foundation models as action priors $\mathbf{a}_N=\{a_k\}_{k=0,1,...,N}$, $N$ generally has a controllable upper limit.

{\bf Text-aware Alignment as Likelihood.} Based on the action priors generated by the action foundation model, the likelihood $L(\ell|\mathbf{v},\mathbf{a})$ aligns the action priors with the given language instruction. As shown in Fig.~\ref{fig:pre_arch}, we leverage pre-trained vision-language foundation model CLIP~\cite{radford2021learning,zhou2022maskclip} to extract vision patch tokens $\mathbf{v}=\{v_i\}$ of the third-view image, text tokens $\ell=\{l_j\}$ and global text embedding $\tilde{l}$ of language instruction. We exploit the text-image alignment property of CLIP to develop text-aware visual features. Specifically, we compute the cosine similarity between each vision patch token $v_i$ and the global text embedding $\tilde{l}$ to get a similarity score $s_i$. These scores then weight the vision tokens as text-aware visual representations $\mathbf{v^\ell}=\{v^\ell_i\}$.
\begin{equation}
\label{eq:text-aware-vision}
    v^\ell_i = s_i \cdot v_i
\end{equation}
To align action priors with task-conditioned vision-language priors, we employ transformer’s attention mechanism~\cite{vaswani2017attention}: $\text{Attention}(Q, K, V)=\text{Softmax}\left({Q K^T}\right) V$, where $Q, K, V$ denote query, key, and value. We concatenate the text-aware vision tokens $v^\ell_i$ with the text tokens $l_j$ to capture the vision-language information. Action poses are encoded via an MLP into action tokens. Cross attention layers then take action tokens as queries, and visual-language features as keys and values, producing $L$ fused action tokens $\mathbf{a}^{vl}_L$. We use rotary position embedding~(RoPE)~\cite{su2024roformer} to encode relative patch position embeddings for query, keys, and values:
\begin{equation}
\begin{aligned}
Q&=\mathrm{RoPE}\left(\mathbf{a}_N\right) \\
K,V&=[\mathrm{RoPE}\left(\mathbf{v^\ell}\right), \ell ] \\
\end{aligned}
\end{equation}
Based on the fused action tokens $\mathbf{a}^{vl}_N$, the likelihood model predicts the probabilities that represent the alignment between the contact poses and the language instruction. 

{\bf Training and Inference.} Given a dataset of triplets $(\mathbf{a},\mathbf{v},\ell)$, the likelihood model is trained using a cross-entropy loss to assign higher probabilities to action proposals consistent with the language instructions. During training, the parameters of the pre-trained foundation models are kept fixed to preserve their generalization capabilities. 
During inference, the prior model $\pi^{p}(\mathbf{a}|\mathbf{v})$ first generates multiple candidate contact poses, followed by the likelihood model $L(\ell|\mathbf{v},\mathbf{a})$ to select a contact pose to execute via motion planning and inverse kinematics.

\subsection{Post-contact Phase}
\label{sec:post-contact}

\begin{figure}[t]
  \centering
  \includegraphics[width=\linewidth]{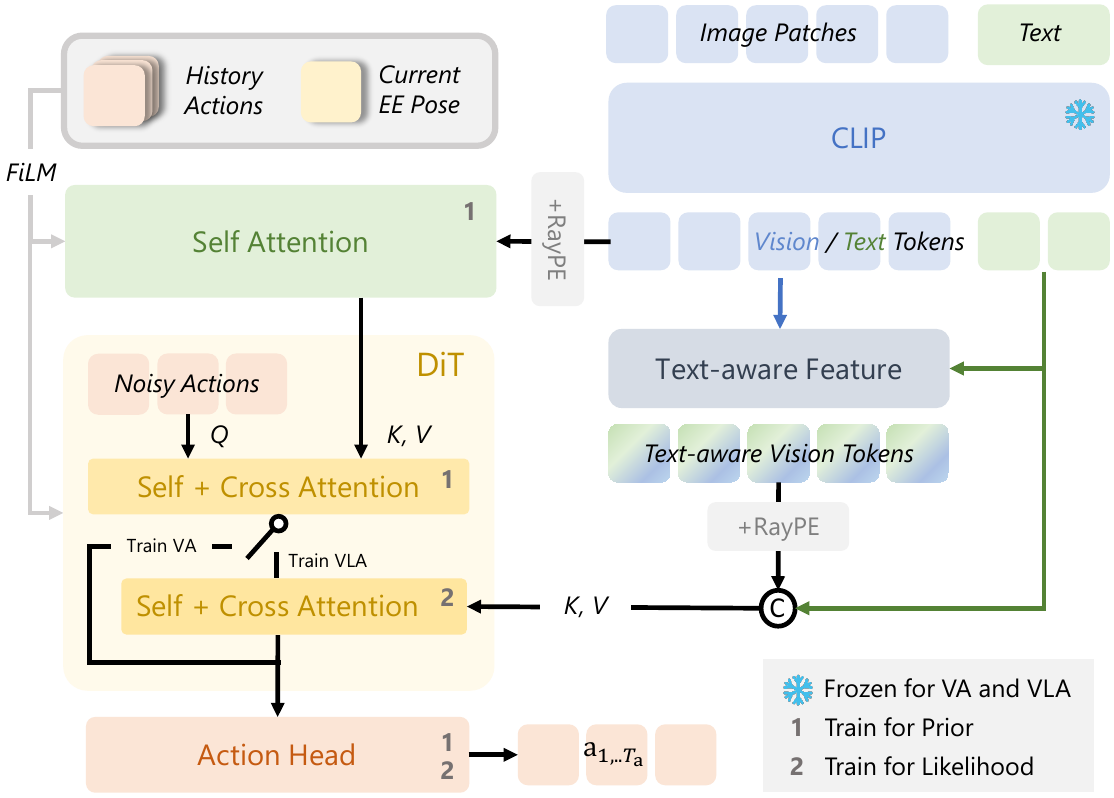}
  \caption{Architecture of post-contact phase. We first train a diffusion-based trajectory generator with vision tokens. Then, we late-inject additional attention layers to condition text tokens and text-aware vision tokens for language alignment. We only tune the additional attention layers and the action head for latent adaptation. }
  \label{fig:post_arch}
\end{figure}

Post-contact skills encompass all manipulation behaviors after contacting objects. To represent general manipulation, the prior $\pi^{p}(\mathbf{a}|\mathbf{v})$ is supposed to predict dense trajectories conditioned on visual observations. In this section, we introduce the prior and likelihood formulations, as well as the training recipe for post-contact phase. Notably, the same modeling approach can also be applied to pre-contact actions.

{\bf Multimodal Trajectory Generator as Prior.} For post-contact actions, the prior $\pi^{p}(\mathbf{a}|\mathbf{v})$ is instantiated as a diffusion-based trajectory generator. Given the visual observation, the prior learns to generate diverse and feasible post-contact trajectories representing possible manipulation behaviors. It aims to capture multimodal action patterns across the visual contexts, serving as action affordances for downstream tasks. Specifically, the architecture follows a vision-conditioned diffusion transformer backbone~(Fig.~\ref{fig:post_arch}). First, the CLIP vision patch tokens of multi-view images are processed by repeated self-attention layers. Plücker raymap based camera pose embedding~(RayPE)~\cite{zhang2024cameras} is applied to all vision tokens. Then, the noisy action tokens are processed by self-attention and cross attention with vision tokens. We use FiLM~\cite{perez2018film} to inject history actions, robot proprioception, and denoise time step into attention layers and feed forward layers. Through this process, we can obtain the fused action tokens $\mathbf{a}^{v}$. Notably, the prior is language-agnostic, which focuses purely on modeling visuomotor distributions. 
\begin{equation}
\begin{aligned}
Q&=\mathrm{RayPE}\left(\mathbf{a}\right) \\
K,V&=\mathrm{RayPE}\left(\mathbf{v}\right) \\
\end{aligned}
\end{equation} 
Given the fused action tokens $\mathbf{a}^{v}$, the action head predicts an action chunk $\mathbf{a}_{1\cdots T_{\mathbf{a}}}$ by recurrent denoising, where $T_\mathbf{a}$ denotes the trajectory horizon.

{\bf Latent Adaptation as Likelihood.} Unlike pre-contact, there are no well-trained models for post-contact that can generate trajectories covering comprehensive feasible motions under diverse visual observations. Therefore, instead of explicitly selecting among action proposals produced by the prior, the likelihood $L(\ell|\mathbf{v},\mathbf{a})$ performs latent adaptation to inject language instruction into the diffusion backbone. As shown in Fig.~\ref{fig:post_arch}, text-aware vision tokens derived as Eqn.~\ref{eq:text-aware-vision} and text tokens are late-injected into the pre-trained diffusion backbone through additional attention layers. Similarly, we use RayPE for text-aware vision tokens. 
\begin{equation}
\begin{aligned}
Q&=\mathrm{RayPE}\left(\mathbf{a}^v\right) \\
K,V&=[\mathrm{RayPE}\left(\mathbf{v^\ell}\right), \ell] \\
\end{aligned}
\end{equation}
This design allows the network to modulate intermediate latent space using language embeddings while preserving the diverse visuomotor representations learned by the prior. During adaptation, only the injected attention layers and the action head are fine-tuned, enabling the model to generate trajectories aware of language instructions.

{\bf Training and Inference.} For post-contact, training contains two stages. In the first stage, the prior is trained on large-scale vision–action pairs using the standard denoising diffusion objective, encouraging the model to generate multimodal action trajectory distributions. In the second stage, we plug additional attention layers into the pre-trained prior model for text-aware token injection. And we only train these layers and the action head on additional paired $(\mathbf{v},\mathbf{a}, \ell)$ data as likelihood, aligning the latent diffusion space with language-conditioned representations. This modular training strategy ensures that the prior maintains general trajectory diversity, while the likelihood introduces language alignment to ground the generated trajectories in language intent. Generally, we split the original datasets into two parts: half for prior training and the rest for likelihood training. This prevents the likelihood from collapsing into an identity mapping by ensuring it is trained on novel $(\mathbf{a},\mathbf{v},\ell)$ tuples, forcing it to rely on language cues rather than merely replicating the prior's knowledge.

\subsection{Pre- and Post-Training Recipe}

Standard VLA policies are typically trained through large-scale pre-training and task-specific post-training. Notably, pre-training is optional, {\it i.e.} a VLA policy can be trained from scratch directly on post-training task-specific data.

{\bf Different Training Recipes.} In our Bayesian formulation, the policy is explicitly decomposed into a vision-action prior and a language-conditioned likelihood. Without pre-training, we simply follow the two-stage procedure described in Sec.~\ref{sec:post-contact}: first train the VA prior with a subset of data, then train the VLA likelihood with the rest. When both pre-training and post-training are considered, the factorization naturally leads to two distinct training recipes:
\begin{itemize}
    \item {\it R1}: Pre-training both the VA prior and the VLA likelihood, while post-training only the VLA likelihood.
    \item {\it R2}: Pre-training only the VA prior, while post-training both the VA prior and the VLA likelihood.
\end{itemize}

{\bf Choice of Training Recipe.} We argue that the likelihood model $L(\ell|\mathbf{v},\mathbf{a})$ requires sufficiently diverse language annotations to avoid collapsing into visual-action shortcuts~(Eqn.~\ref{eq:loss}). Therefore, selecting between {\it R1} and {\it R2} depends critically on the amount and diversity of language instructions available during post-training.

{\it When to use R1.} For small-scale post-training datasets where language annotation diversity is extremely limited, updating the VA prior during post-training may lead to language neglect. In these cases, we adopt {\it R1} to perform language alignment during pre-training and apply only likelihood fine-tuning during post-training. This preserves the capability in pre-trained VA prior while enabling language-conditioned likelihood adaptation. Such cases include most real-world datasets or some simulation benchmarks, {\it e.g.} LIBERO~\cite{liu2023libero}.

{\it When to use R2.} If the post-training dataset is sufficiently large, {\it e.g.}, $\geq1k$ trajectories, and exhibits a substantial vision–action distribution shift relative to pre-training, then continuing to update the VA prior during post-training becomes beneficial. In this case, the language annotations must be sufficiently diverse to support likelihood fine-tuning. Such conditions are typically easier to satisfy in simulation, where both visual scenes and language prompts can be varied broadly.

\subsection{Implementation Details}

\begin{figure}[t]
  \centering
  \includegraphics[width=\linewidth]{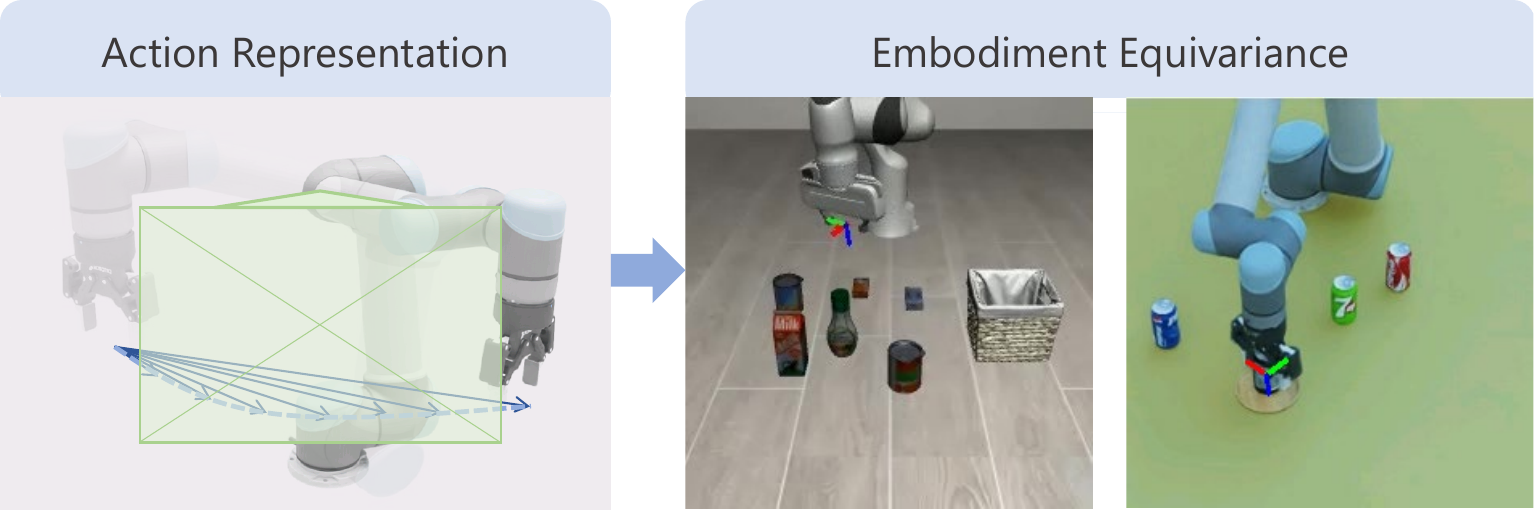}
  \caption{Our action representation is formulated as relative end-effector poses projected in the camera frame. Such representation disentangles the embodiment-relative information to achieve embodiment equivariance.}
  \vspace{-0.1cm}
  \label{fig:action-repre}
\end{figure}

{\bf Embodiment Equivariant Action Representation.} {To leverage the cross-embodiment information, we disentangle the embodiment-dependent information in policy to achieve embodiment equivariance. For pre-contact phase, the grasp poses are represented as end-effector poses projected in camera frame. For post-contact phase, we employ the action representation in ~\cite{chen2025e2vla}. As shown in Fig.~\ref{fig:action-repre}, the action is represented as the relative motion of end-effector projected in camera frame. Specifically, denote the action representation in robot data as $\mathbf{a}^*$, it can be formulated as:}
\begin{equation}
    \mathbf{a}^* = {}^c_{e^*}\mathbf{T}{}^e_{c}\mathbf{T}
\end{equation}
where $c$ represents camera frame, $e$ represents end-effector frame. In this way, the policy prediction is equivariant to different embodiment configurations. Then, the predicted action can be transformed to the base frame for robot control with camera extrinsics by calibration. To further mitigate the influence from extrinsic calibration error, we refine the action representation to:
\begin{equation}
\mathbf{a}^* = \left[\begin{array}{cc}
{ }_{e_*}^c \mathbf{R}{ }_c^e \mathbf{R} & { }_e^c \mathbf{R} {}_{e_*}^e\mathbf{t} \\
\mathbf{0} & 1
\end{array}\right]
\end{equation}

{\bf $\mathrm{SE}(3)$ Action.} All the actions are defined on the $\mathrm{SE}(3)$ manifold, where rotations are represented using the 6D continuous representation~\cite{zhou2019rot6d}. For post-contact phase, we further predict the normalized gripper width, where $-1$ denotes fully closed and $1$ denotes fully open. In total, the action vector has 10 dimensions: 3 for translation, 6 for rotation, and 1 for gripper control. 

{\bf Network Architecture.} We use the checkpoint of CLIP ViT-L/14 and adopt MaskCLIP technique~\cite{zhou2022maskclip} to extract vision patch tokens. Separate projection MLP layers are applied to transform vision tokens, text-aware vision tokens, and text tokens into the model’s embedding space. Note that the text-aware mechanism operates on the original CLIP vision and text tokens. For the pre-contact phase, we use four cross attention blocks. For post-contact phase, we employ four self attention blocks to process vision tokens. The diffusion transformer contains four self$\&$cross attention blocks for the prior, and an additional four self$\&$cross attention blocks for the likelihood. We set the trajectory horizon to $T_\mathbf{a}=16$ in simulation and $T_\mathbf{a}=32$ in the real world. 

{\bf Training Settings.} We adopt the AdamW optimizer with a learning rate of 1e-4 and a weight decay of 1e-2. The models are trained with a batch size of 32 using mixed precision~(float32 and bfloat16). The prior is trained for 400k iterations, and the likelihood for 200k iterations. During training, we employ a 100-step DDPM schedule, while for inference we use a 20-step DDIM sampler.

\section{Information-theoretic Results}
\label{sec:inform-theory}
In this section, information theory is employed to validate that language may be ignored in the resultant VLA using general training method. We also show that our two-stage prior and likelihood training can be superior. 
Formally, minimizing the VLA objective corresponds to minimizing the conditional negative log-likelihood of actions given vision and language. This objective is closely connected to conditional entropy:
\begin{equation}
      \min -\mathbb{E} [\log\pi(\mathbf{a}|\mathbf{v},\ell)] \approx \mathcal{L}_\mathrm{VLA} = H(\mathbf{a}|\mathbf{v},\ell)
\end{equation}
Thus, the optimal VLA model approaches the minimum achievable loss $\mathcal{L}_\mathrm{VLA}$, which equals the conditional entropy $H(\mathbf{a}|\mathbf{v},\ell)$. When training converges, the negative log-likelihood objective is expected to approximate this entropy lower bound.

{\bf Dataset Diversity Bias.} In most VLA datasets, there exists a bias in the diversity between language modality and the vision-action modality. Typically, a data trajectory consisting of many visual frames is associated with a single language prompt. Although two trajectories with different prompts may share a small number of visually similar frames, such overlaps occur only rarely. As a result, for most visual observations, the corresponding language instruction is nearly deterministic. Formally, denoting $P(\ell|\mathbf{v})$ as the probability of language prompt conditioned on the visual frame, we state the assumption that $P(\ell|\mathbf{v})$ is effectively one-hot for the majority of frames. In other words, the language can often be inferred directly from the visual observation. Consequently, the conditional entropy is upper-bounded by a small constant, {\it i.e.}, $H(\ell|\mathbf{v}) \leq \epsilon$. For example, if a drawer is visible in a frame, the associated instruction is likely to involve opening or closing the drawer. Such bias is prevalent in many existing embodied datasets and implies that language provides limited additional information beyond what is already encoded in the visual input.
\subsection{Shortcut Learning}
Given the assumption, we show that general VLA training admits a shortcut solution that largely ignores language. We begin by defining the conditional mutual information $I$ as: 
\begin{equation}
\begin{aligned}
    I(\mathbf{a};\ell|\mathbf{v}) 
    & = H(\mathbf{a}|\mathbf{v}) - H(\mathbf{a}|\mathbf{v},\ell) \\
    & = H(\ell|\mathbf{v}) - H(\ell|\mathbf{v},\mathbf{a})
\end{aligned}
\end{equation}
where $H(\cdot)\geq0$. Based on information theory and the above assumption, we have:
\begin{equation}
    0 \leq I(\mathbf{a};\ell|\mathbf{v}) \leq H(\ell|\mathbf{v})  \leq \epsilon
\end{equation}
leading to
\begin{equation}
    0 \leq  H(\mathbf{a}|\mathbf{v}) - H(\mathbf{a}|\mathbf{v},\ell)   \leq \epsilon
    \label{avavl}
\end{equation}
which means that the inclusion of the language prompt makes limited contribution to the prediction of action. 

Considering an alternative model trained with the objective 
\begin{equation}
    \min -\mathbb{E}[\log\pi(\mathbf{a}|\mathbf{v})]
    \label{vamodel}
\end{equation}
whose optimal lower bound $\mathcal{L}_\mathrm{VA}$ is the conditional entropy $H(\mathbf{a}|\mathbf{v})$. Then we have
\begin{equation}
    \mathcal{L}_\mathrm{VLA} \le \mathcal{L}_\mathrm{VA} \leq \mathcal{L}_\mathrm{VLA} +\epsilon
    \label{eq:loss}
\end{equation}
It means that a vision-only policy can achieve only slightly larger loss than the desired VLA policy. This vision-only policy is simpler than the vision-language policy, as the language input is ignored. Therefore, there is a \textit{shortcut} for VLA learning, which causes the final learned model to be $\pi(\mathbf{a}|\mathbf{v})$. This theoretical result explains that the bias in language diversity in the dataset makes the existing VLA weak in OOD instruction following. 

\subsection{Two-stage Conditioning}

Based on the Bayesian factorization, our training is actually two-stage. In the first stage, we learn a vision-action expert, which follows Eqn.~\ref{vamodel}. Note that the prior model $\pi^p(\mathbf{a}|\mathbf{v})$ in Eqn.~\ref{eq:bayesian} does not take in language prompt, so it is exactly the desired policy, not a shortcut. Denote the learned parameter as $\theta_\mathrm{VA}$, we have
\begin{equation}
    \mathcal{L}_\mathrm{VA} \approx \min -\mathbb{E}[\log \pi^p_{\theta_\mathrm{VA}}(\mathbf{a}|\mathbf{v})] 
\end{equation}
Denote the intermediate embedding of $\pi^p_{\theta_\mathrm{VA}}(\mathbf{a}|\mathbf{v})$ as $f_{\theta_\mathrm{VA}}(\mathbf{v})$, we have the policy in the second stage of language prompt injection to specify an action by
\begin{equation}
    \mathcal{L}_\mathrm{VLA} \leq \min -\mathbb{E}[\log\pi(\mathbf{a}|f_{\theta_\mathrm{VA}}(\mathbf{v}),\ell)] \leq \mathcal{L}_\mathrm{VA}
    \label{stage2}
\end{equation}
where $\theta_\mathrm{VA}$ is frozen at the result of the first stage, leaving only the learnable parameters from language prompt. If the language information is entirely unused, then the second equation of Eqn.~\ref{stage2} exists. Since the loss can be further optimized, the gradient enforces the network to utilize the language information for action generation. As the policy now conditions on embedding of the image is used rather than the original one, the minimized loss $\min -\mathbb{E}[\log\pi(\mathbf{a}|f_{\theta_\mathrm{VA}}(\mathbf{v}),\ell)]$ may be larger than the lower bound $\mathcal{L}_\mathrm{VLA}$. At last, we have a resultant policy with a loss lower than the vision-only shortcut model, and effectively taking the language prompt.

\subsection{Model Capacity}

Since the embedding of vision is utilized for injection, following data processing inequality~\cite{CMU-10704}, we have
\begin{equation}
   I(\mathbf{a};\ell|f_{\theta_\mathrm{VA}}(\mathbf{v})) \le I(\mathbf{a};\ell|\mathbf{v}) \le \epsilon
   \label{mi1}
\end{equation}
which shows a small mutual information between action and language given the vision embedding. Then, given the mutual information between action and vision-language $I(\mathbf{a};\mathbf{v},\ell)$, we have
\begin{equation}
    I(\mathbf{a};\mathbf{v},\ell) = H(\mathbf{a})-H(\mathbf{a}|\mathbf{v},\ell)
\end{equation}
According to Eqn.~\ref{avavl}, we have 
\begin{equation}
    I(\mathbf{a};\mathbf{v},\ell) \approx  H(\mathbf{a})-H(\mathbf{a}|\mathbf{v}) = I(\mathbf{a};\mathbf{v})
    \label{mi2}
\end{equation}
Note that
\begin{equation}
   I(\mathbf{a};\mathbf{v},\ell) = I(\mathbf{a};\mathbf{v}) +I(\mathbf{a};\ell|\mathbf{v})
\end{equation}
combining Eqn.~\ref{mi1} and Eqn.~\ref{mi2}, we have
\begin{equation}
    I(\mathbf{a};\ell|f_{\theta_\mathrm{VA}}(\mathbf{v})) \ll I(\mathbf{a};\mathbf{v},\ell)
    \label{size}
\end{equation}
which indicates that the total information for the VLA policy is much larger than the information for language conditioning when there is a bias in the dataset.

In information theory, the model size is related to the mutual information between the model input and output. Therefore, based on Eqn.~\ref{size}, we know that the model size required for the original VLA policy is much larger than that for language conditioning. So, this conditioning network can be small but specialized for language injecting, making high efficacy to reserve generalization during post-training, even with small data.

\begin{figure*}[t]
  \centering
  \includegraphics[width=\linewidth]{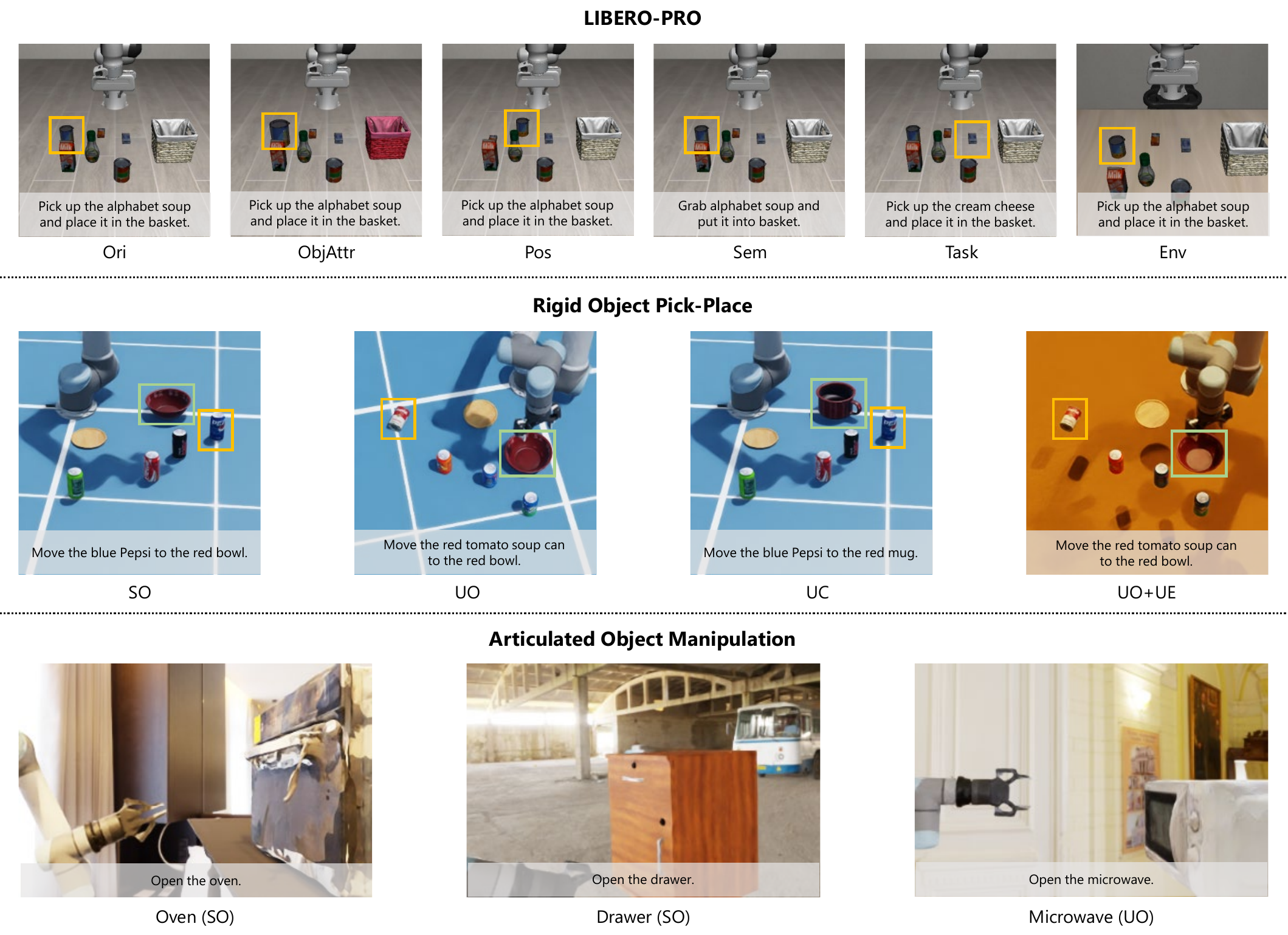}
  \caption{Our simulation benchmarks contain the commonly-used LIBERO(-PRO) and two self-built ones, including rigid object pick-place and articulated object manipulation. Each benchmark demonstrates both seen and unseen settings. For LIBERO, besides the original setting~(Ori), we evaluate the policy under perturbations on object attributes~(ObjAttr), initial positions~(Pos), language semantic expressions~(Sem), task instructions~(Task), and environments~(Env). For rigid object pick-place, we test four settings: Seen Object~(SO), Unseen Object~(UO), Unseen Container~(UC), and Unseen Object+Unseen Environment~(UO+UE). For articulated object manipulation, we test two seen object tasks of oven opening~(Oven) and drawer opening~(Drawer), and one unseen object task for microwave opening~(Micro).}
  \label{fig:sim_bench}
  \vspace{-0.2cm}
\end{figure*}
\section{Simulation Experiments}
\label{sec:exp}
In this section, we carry out a series of simulation experiments to evaluate our policy. The goals of the experiments are: 1) to test the effectiveness of Bayesian factorization in VLA; 2) to evaluate the zero-shot or few-shot generalization performance of our policy on diverse dimensions; 3) to validate the architecture designs of pre-contact and post-contact.  

\subsection{Benchmarks}
In our simulation experiments, we test the models on commonly-used LIBERO benchmarks~\cite{liu2023libero,zhou2025liberopro}. To test policies on diverse language instructions and unseen settings, we further design two simulation benchmarks that include rigid object pick-place and articulated object manipulation tasks based on previous work~\cite{chen2025e2vla,yang2025disambiguate}. Fig.~\ref{fig:sim_bench} shows an overview of the three simulation benchmarks.

{\bf LIBERO and LIBERO-PRO.} We use the LIBERO-Object task suite~\cite{liu2023libero}, which maintains identical scene layouts but introduces variations in object types, evaluating the policy’s capacity to distinguish among different object instances. The task suite contains 10 tasks with 50 demonstrations each, leading to a total of 500 trajectories. Since the original datasets lack camera parameters, we follow OpenVLA~\cite{kim2024openvla} to replay the ground truth actions and record this information. Failure demonstrations are removed before training. Also, to train the prior for pre-contact, we deploy AnyGrasp~\cite{fang2023anygrasp} to generate grasp proposals and execute the one that satisfies the language instructions. To further test out-of-distribution generalization, we adopt the LIBERO-PRO protocol~\cite{zhou2025liberopro}, which introduces reasonable perturbations across five dimensions: manipulated object attributes~(ObjAttr), initial positions~(Pos), language semantic expressions~(Sem), task instructions~(Task), and environments~(Env). 

{\bf Rigid Object Pick-Place.} We develop a benchmark in IssacSim~\cite{NVIDIA_Isaac_Sim} for diverse object pick-place tasks using a UR5 robot arm. For each task, we randomly sample 4 of 25 objects that cover diverse colors, shapes, sizes, and categories, with a randomly selected object to pick. Each task randomly selects a place target from a yellow plate and a red bowl. We collect 10k pick-place trajectories across 500 scenarios for multi-task training, each with randomized camera viewpoints and object layouts. To test the generalization, we construct four suites: Seen Object~(SO), Unseen Object~(UO), Unseen Container~(UC), and Unseen Object+Unseen Environment~(UO+UE). For each task in SO, we use the seen pick objects and containers, but with randomly sampled language instructions, camera viewpoints and object layouts. For each task in UO, we randomly sample 4 of 10 unseen objects covering unseen colors, shapes, sizes, and categories, but with seen place target. This task suite test generalization of manipulated object categories for pre-contact actions. For each task in UC, we use the seen objects, but with unseen container. This task suite test generalization in place target object categories for post-contact actions. For UO+UE, we further apply unseen environments including backgrounds and lights for UO, to evaluate the robustness under diverse environments. Notably, language instructions, camera viewpoints, and object layouts are randomly sampled through all testing.

{\bf Articulated Object Manipulation.} Following \cite{yang2025disambiguate}, we test the policies on articulated object manipulation tasks, which involve richer motion patterns and more flexible trajectories compared to object pick-place. We collect training data in IssacLab~\cite{NVIDIA_Isaac_Lab} for two tasks: one oven opening task, and one drawer opening task. Using the demonstration generation procedure from \cite{yang2025disambiguate}, we obtain 2k trajectories for each tasks~(Oven and Drawer). For each rollout during data collection and testing, we apply extensive randomization~(object and robot poses, lighting conditions, and textures) to diversify the scenarios. To assess the generalization, we additionally introduce an unseen articulated object manipulation task to open the microwave~(Micro). This task requires handling novel motion constraints that differ from those in the seen tasks, providing a challenging out-of-distribution evaluation.

\subsection{Dataset Exploratory Analysis}

\begin{table}[t]
\centering
\caption{Dataset Diversity Bias Validation.}
\label{tab:data}
\begin{tabular}{lcccc}
\toprule
Dataset & \# Lang. & $H(\ell|\mathbf{v})$  \\
\midrule
LIBERO & 10 & 1.6$\times$10$^{-9}$ \\
Rigid Object Pick-Place & 75 & 7.2$\times$10$^{-2}$ \\
\bottomrule
\end{tabular}
\vspace{-0.1cm}
\end{table}

{\bf Dataset Bias Validation.} Our first experiment validates the assumption of dataset diversity bias discussed in Sec.~\ref{sec:inform-theory}. For the training dataset of each benchmark, we train a classifier using cross-entropy loss to predict the language instruction based on a randomly selected visual frame from a trajectory. We report the conditional entropy $H(\ell|\mathbf{v})$ of the prediction results and the number of language instructions in Table~\ref{tab:data}. It is shown that LIBERO exhibits extremely low $H(\ell|\mathbf{v})$, indicating that visual cues alone suffice to infer the language instruction. Such a bias in the dataset may encourage policies to rely on visual shortcuts, 
neglecting proper language grounding during action generation. In contrast, our self-designed rigid object pick-place dataset demonstrates substantially higher conditional entropy for language prediction, highlighting the necessity for policies to ground language for effective action generation.

{\bf Deployment of Two Training Recipes.} In general, we pre-train VA prior with DROID~\cite{khazatsky2024droid} dataset, which contains 78,544 trajectories of tabletop manipulation with Franka. LIBERO involves the same embodiment as this pre-training. Besides, Table~\ref{tab:data} shows that the language diversity is limited. Therefore, we deploy the training recipe {\it R1}. Specifically, we pre-train the VA prior on DROID~\cite{khazatsky2024droid} and simulation datasets~(including rigid object pick-place as well as articulated object manipulation), and pre-train the VLA likelihood on these simulation datasets. Then, we fine-tune the VLA likelihood with the LIBERO ones. In contrast, for the rigid object pick-place and articulated object manipulation, the dataset scale is relatively larger and the visual domain is distinct from that of pre-training, {\it e.g.} different embodiment of UR5. To this end, we deploy {\it R2}, we only pre-train the VA prior on DROID, followed by fine-tuning of the VA prior and VLA likelihood on these simulation datasets.

\subsection{Baselines}
Our baselines include diffusion policies that are trained from scratch, and start-of-the-art VLA policies that are pre-trained on large-scale datasets and fine-tuned on benchmark datasets. All baselines learn a unified policy that tightly entangles action generation with language grounding.

{\bf 3D Diffuser}~\cite{ke20243d} is a language-conditioned diffusion policy that incorporates multi-view RGB-D images. It utilizes pre-trained CLIP~\cite{radford2021learning} to extract multi-view features, followed by a 3D denosing transformer for action generation.

{\bf OpenVLA}~\cite{kim2024openvla} is built on Prismatic
7B~\cite{anschutz2023prismatic} VLM backbone, with large-scale pre-training on diverse datasets, including OXE~\cite{openxembodiment}. To test on specific benchmark, we deploy the OpenVLA-OFT fine-tuning recipe with continuous action representation~\cite{kim2025oft}. 

{\bf $\pi_0$}~\cite{black2024pi0} combines PaliGemma~\cite{beyer2024paligemma} and a diffusion-based action expert, which are pre-trained on a subset of OXE and the $\pi$ dataset, and fine-tuned on the DROID~\cite{khazatsky2024droid} dataset. 

{\bf $\pi_{0.5}$}~\cite{black2025pi_ohfive} is the latest state-of-the-art VLA model. Compared to $\pi_0$, $\pi_{0.5}$ uses a hybrid training procedure, with additional discrete tokenized pre-training on a broad range of robot and non-robot data.

\subsection{Results on LIBERO and LIBERO-PRO}
\begin{table}[t]
\centering
\caption{Simulation Results on LIBERO and LIBERO-PRO.}
\label{tab:libero}
\begin{tabular}{lcccccc}
\toprule
Method & Ori & ObjAttr & Pos & Sem & Task & Env \\
\midrule
OpenVLA & 98 & {\bf 98} & 0 & {\bf 98} & 0 & 0 \\
$\pi_0$    & {\bf 99} & 94 & 0 & 90 & 0 & 29 \\
$\pi_{0.5}$   & 98 & {\bf 98} & 17 & 96 & 1 & 73 \\
\midrule
BayesVLA & {\bf 99} & 85 & 15 & {\bf 98} & {\bf 10} & 1 \\
+ R1 & {\bf 99} & 96 & {\bf 21} & {\bf 98} & {\bf 10} & {\bf 81}\\
\bottomrule 
\end{tabular}
\end{table}

{\bf Original LIBERO Settings.} Results in Table~\ref{tab:libero} show that all the models achieve strong performance on the original LIBERO-Object tasks, owing to the limited task variations and sufficient demonstrations. Nevertheless, BayesVLA matches or slightly exceeds baseline methods without any external pre-training. This suggests that the Bayesian factorization effectively captures the task structure from scratch, allowing the model to efficiently learn both visuomotor priors and language-conditioned likelihood in data-sufficient scenarios.

{\bf Zero-shot Generalization.} When introducing systematic perturbations across object attributes, initial positions, semantics, and environments as in LIBERO-PRO~\cite{zhou2025liberopro}, previous VLAs show strong biases toward seen configurations~(Table~\ref{tab:libero}). In particular, these policies are sensitive to position perturbation and changes in task instructions. Simply swapping the target object with another one causes substantial performance degradation, revealing their limited semantic grounding capability. Likewise, the performance collapses a different target object within the original scenario. Such poor language following indicates that these policies rely heavily on visual cues for action generation, rather than genuinely grounding the language instructions. This behavior can be attributed to the extremely small conditional entropy $H(\ell|\mathbf{v})$ in the LIBERO training dataset, which encourages the policy to exploit shortcuts that ignore the language. By decomposing the policy into a vision-action prior and a language-conditioned likelihood, BayesVLA with {\it R1} pre-training strengthens language grounding for action generation, achieving improved robustness under position and instruction perturbations. However, due to the limited language diversity in the training dataset, the overall success rate remains low. For environment variation, BayesVLA trained from scratch performs poorly because of the shifted relative positions between the robot and the table, whereas {\it R1} pre-training effectively mitigates this issue. Finally, all policies achieve good performance under unseen object attributes and semantics, as these variations do not require novel action trajectories and introduce only minor differences in the visual inputs.

\subsection{Results on Rigid Object Pick-Place.}
\begin{table}[t]
\centering
\caption{Simulation Results on Rigid Object Pick-Place.}
\label{tab:pick-place}
\begin{tabular}{lcccc}
\toprule
{Method} & {SO} & {UO} & {UC} & {UO+UE} \\
\midrule
3D Diffuser & 38 & 10 & 12 & 4 \\
$\pi_0$ & 26 & 10 & 10 & 10 \\
$\pi_{0.5}$ & 48 & 38 & 40 & 32 \\
\midrule
BayesVLA & {\bf 84} & {\bf 56} & {\bf 60} & {\bf 46} \\
+ R2 & {\bf 84} & {\bf 56} & {\bf 60} & {\bf 46} \\
\bottomrule
\end{tabular}
\vspace{-0.2cm}
\end{table}

Table~\ref{tab:pick-place} summarizes the simulation results on the rigid object pick-place benchmark. In general, BayesVLA consistently outperforms all baselines across seen and unseen configurations.

{\bf Seen Objects.} For this setting, all objects appear during training, but test-time language instructions are randomly sampled, calling for language following ability. 3D Diffuser can generate reasonable trajectories but frequently fails to pick up the target object. Even with large-scale pre-training, $\pi_0$ often selects the wrong object due to weak language grounding. This arises because the conditional entropy $H(\ell|\mathbf{v})$ remains low, resulting in unexpected ignorance of language following. It is worth noting that 3D Diffuser outperforms $\pi_0$, highlighting the benefit of 3D visual information even without any pre-training. By co-training on both reasoning and robot data, $\pi_{0.5}$ improves instruction following. However, its dominant failure mode becomes grasp misalignment, leading to unsuccessful pickups despite understanding the instruction. Instead, BayesVLA leverages a vision-action prior to generate plausible pre-contact actions, while the language-conditioned likelihood enforces consistency with the task instruction. This factorization allows BayesVLA to acquire stronger visuomotor skills without compromising language grounding.

{\bf Zero-shot Generalization.} We evaluate policies under three generalization settings: unseen objects~(UO), unseen containers~(UC), and combined unseen environments~(UO+UE), all of which require interpreting previously unseen language instructions. Overall performance drops compared to the seen-object setting, but the gaps remain smaller than those observed on the LIBERO benchmark. This is because this benchmark exhibits a higher conditional entropy, $H(\ell|\mathbf{v}) = 7.2 \times 10^{-2} \textgreater 1.6 \times 10^{-9}$, making language grounding essential and thereby reducing the generalization discrepancy. Both 3D Diffuser and $\pi_0$ consistently fail to ground language on novel object categories, with 3D Diffuser even struggling to generate trajectories directed toward any object in the scene. These models tend to overfit to visual shortcuts and thus cannot generalize under distribution shifts. $\pi_{0.5}$ can produce reasonable grasping actions in the UO and UC settings. However, such performance drops obviously once unseen environments are introduced. Notably, even without {\it R2} pre-training, BayesVLA achieves the best performance across all unseen settings. With sufficient training demonstrations~(10k), the VA prior captures plausible action distributions, while the VLA likelihood inherits the generalization from pre-trained foundation models and effectively aligns action priors with language instructions. Furthermore, due to the substantial embodiment gap between Franka~(used during pre-training) and UR5~(used in evaluation), the {\it R2} pre-training contributes limited benefit. 

\begin{table}[t]
\centering
\caption{One-shot Adaptation on Rigid Object Pick-Place.}
\label{tab:1shot-pick-place}
\begin{tabular}{lcccc}
\toprule
{Method} & {UO} & {UC} & {UO+UE} \\
\midrule
$\pi_0$ & 58 & 82 & 24\\
$\pi_{0.5}$ & {\bf 82} & 82 & -- \\
\midrule
BayesVLA & 78 & {\bf 88} & 62 \\
+ R2 & 80 & {\bf 88} & {\bf 66} \\
\bottomrule
\end{tabular}
\vspace{-0.2cm}
\end{table}

{\bf One-shot Adaptation.} For each task in each unseen setting, we collect a single demonstration for further policy adaptation and report results in Table~\ref{tab:1shot-pick-place}. After adaptation, $\pi_0$ achieves consistent improvements across all generalization dimensions, indicating solid domain adaptation capability. $\pi_{0.5}$ also shows notable gains in UO and UC settings. However, we observe that $\pi_{0.5}$ struggles to fit the UO+UE one-shot data. 
This may be because the environments in the one-shot data differ significantly from those in the original dataset. The pre-trained policy overfits to the original distribution and fails to jointly optimize language alignment and action generation in the coupled feature space on the new dataset. As a result, early fine-tuning steps can trigger gradient explosion, causing the policy collapse. Conversely, BayesVLA exhibits strong one-shot adaptation capability in both pre-contact and post-contact phases. Even under the challenging UO+UE setting, it still achieves an improvement of approximately $20\%$. This suggests that the VA prior captures generalizable action priors, and the VLA likelihood learns adaptive semantic alignment without overwriting the pre-trained VA priors, enabling the policy to adjust to novel scenarios without retraining the full model. Moreover, {\it R2} pre-training further enhances the one-shot adaptation performance.

\subsection{Results on Articulated Object Manipulation}
\begin{table}[t]
\centering
\caption{Simulation Results on Articulated Object Manipulation.}
\label{tab:articulated}
\begin{tabular}{lcccc}
\toprule
\multirow{2}{*}{Method} & \multicolumn{2}{c}{SO} & \multicolumn{2}{c}{UO} \\
\cmidrule(lr){2-3} \cmidrule(lr){4-5}
 & Oven & Drawer & Micro & Micro~(10shot) \\
\midrule
3D Diffuser & 86 & 48 & 0 & 16 \\
$\pi_0$ & 85 & 1 & 7 & 17\\
$\pi_{0.5}$ & {\bf 98} & 24 & 10 & 32 \\
\midrule
BayesVLA & 92 & {\bf 86} & {\bf 38} & {\bf 74} \\
+ R2 & 92 & {\bf 86} & {\bf 38} & {\bf 74} \\
\bottomrule
\end{tabular}
\vspace{-0.1cm}
\end{table}

Table~\ref{tab:articulated} reports the results on the articulated object manipulation benchmark, including oven, drawer, and microwave opening tasks.

{\bf Seen Objects.} For oven opening, all policies achieve strong performance across diverse environments. However, for drawer opening, the success rate of all baselines drops substantially. For instance, $\pi_0$ and $\pi_{0.5}$ struggle to grasp the drawer handle, which is considerably smaller and harder to localize than the oven handle. By leveraging 3D information, 3D Diffuser achieves a much better success rate. BayesVLA benefits from the VA prior that provides favorable action priors and thereby maintains robust performance on drawer opening tasks. 

{\bf Zero/Few-shot Generalization.} For the unseen microwave task, all baselines show low performance~($\leq 10\%$). consistently failing to localize the microwave handle and execute effective motion under the distinct articulation axis and motion constraints. Notably, even in rare successful cases, the policies often push the door open directly, rather than following the correct motion constraints. In contrast, BayesVLA benefits from the generalizable pre-contact VA prior, which enables more reliable handle localization. Therefore, it can achieve a 38$\%$ success rate. With a small amount of adaptation data~(10-shot setting), the success rate further increases to 74$\%$, validating that BayesVLA can adapt to novel kinematic structures with minimal fine-tuning. This underscores the effectiveness of our latent adaptation mechanism, which supports efficient likelihood fine-tuning in latent space.

\subsection{Ablation Studies}
To further understand the contribution of each component, we conduct extensive ablation studies to verify: 1) the advantage of our prior-likelihood factorization; 2) the effectiveness of our two phases of manipulation~(pre-contact $\&$ post-contact); 3) the impact of pre-training; 4) the design choices of likelihood in pre-contact and post-contact.

\begin{figure}[t]
  \centering
  \includegraphics[width=\linewidth]{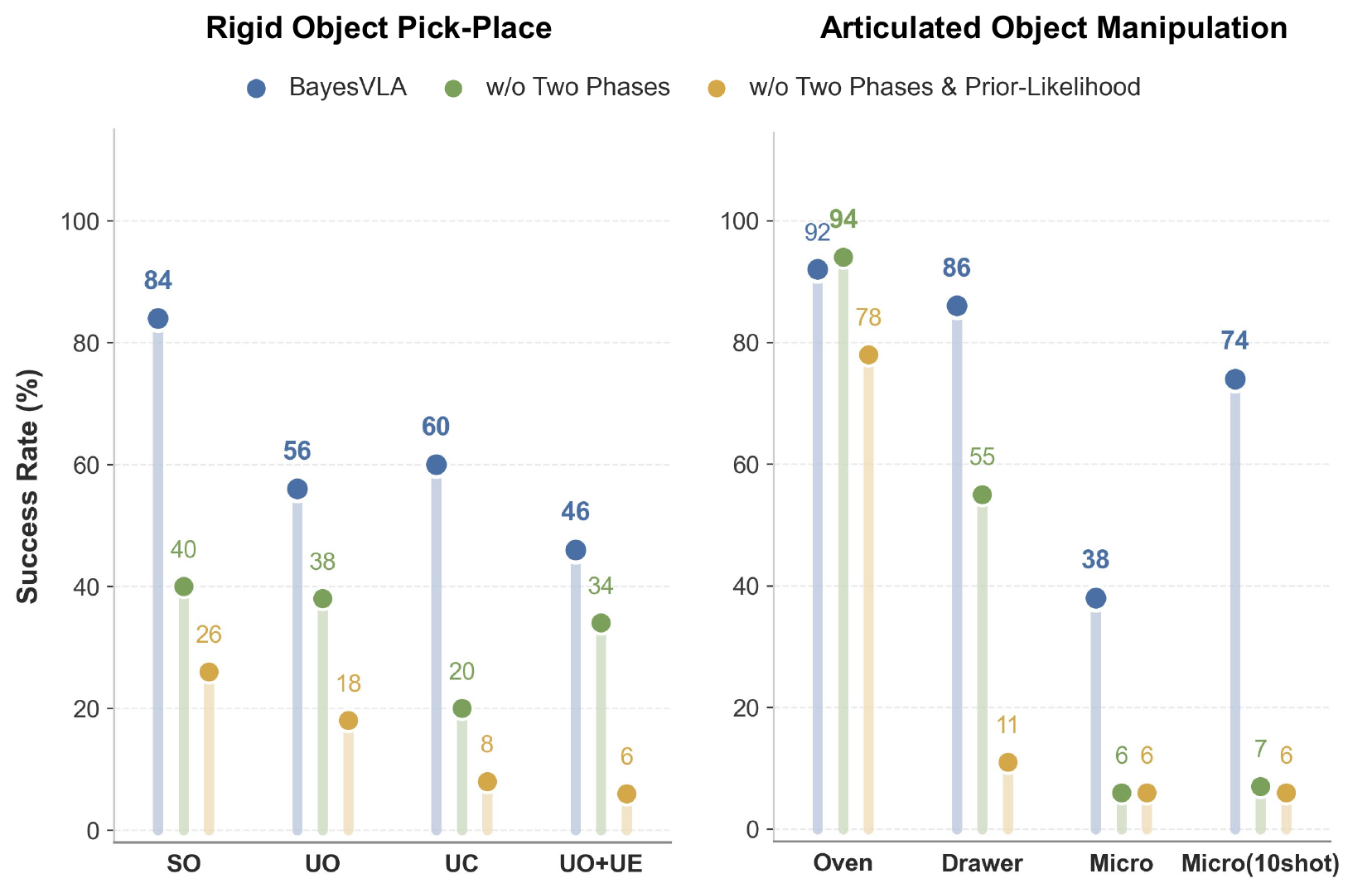}
  \caption{Ablation of two-phase design~(pre-contact $\&$ post-contact) and prior-likelihood factorization.}
  \label{fig:two-phase}
  \vspace{-0.1cm}
\end{figure}

{\bf Pre-/Post-contact Phases.} We compare our method against a variant that does not distinguish between the pre-contact and post-contact phases~(Fig.~\ref{fig:two-phase}). Specifically, this variant applies the post-contact prior and likelihood architecture with the same two-stage training procedure, and is trained directly on dense trajectories covering both pre-contact and post-contact phases. Therefore, it is unable to leverage pre-trained priors for pre-contact poses. The results show a substantial performance degradation across all scenarios, especially on novel articulated object manipulation tasks. Without pre-contact priors, the policy struggles to accurately localize novel objects and predict feasible contact poses even with 10-shot adaptation. It can only generalize to some objects with similar contact configurations in pick-place tasks. By explicitly decomposing the policy into two phases, we can better capture the structure of object manipulation: before contact, the crucial step is to predict a feasible contact pose, which allows us to exploit powerful pre-trained models. Since contact-pose data are easier to collect and typically available in larger quantities, such pre-trained models provide stronger robustness and generalization. 
Combining results in Tab.~\ref{tab:pick-place} and Tab.~\ref{tab:articulated}, this variant outperforms $\pi_0$, verifying the effectiveness of Bayesian factorization, and suggesting that simply enlarging the pre-training dataset is insufficient to improve generalization. While its performance is slightly lower than $\pi_{0.5}$, it attains results comparable to a far more expensive co-training approach using both VLA and reasoning datasets.

{\bf Prior-Likelihood Factorization.} Building on the above experiment, we further remove the prior–likelihood factorization and adopt the standard VLA training paradigm. As shown in Fig.~\ref{fig:two-phase}, this leads to a substantial performance drop across all scenarios. Even in the seen-object setting, the model fails to reliably follow language instructions or localize the correct target object. These observations support our hypothesis that conventional VLA training encourages shortcut learning based on visual cues, resulting in poor language grounding and limited generalization. Instead, by factorizing the policy as prior and likelihood, we mitigate this issue through a principled framework, yielding better generalization performance. Compared with $\pi_0$ in Tab.~\ref{tab:pick-place} and Tab.~\ref{tab:articulated}, this variant shows only slight performance degradation. This again suggests that merely increasing the amount of VLA data without increasing the language diversity does not alleviate the underlying shortcut learning problem.

\begin{figure}[t]
  \centering
  \includegraphics[width=\linewidth]{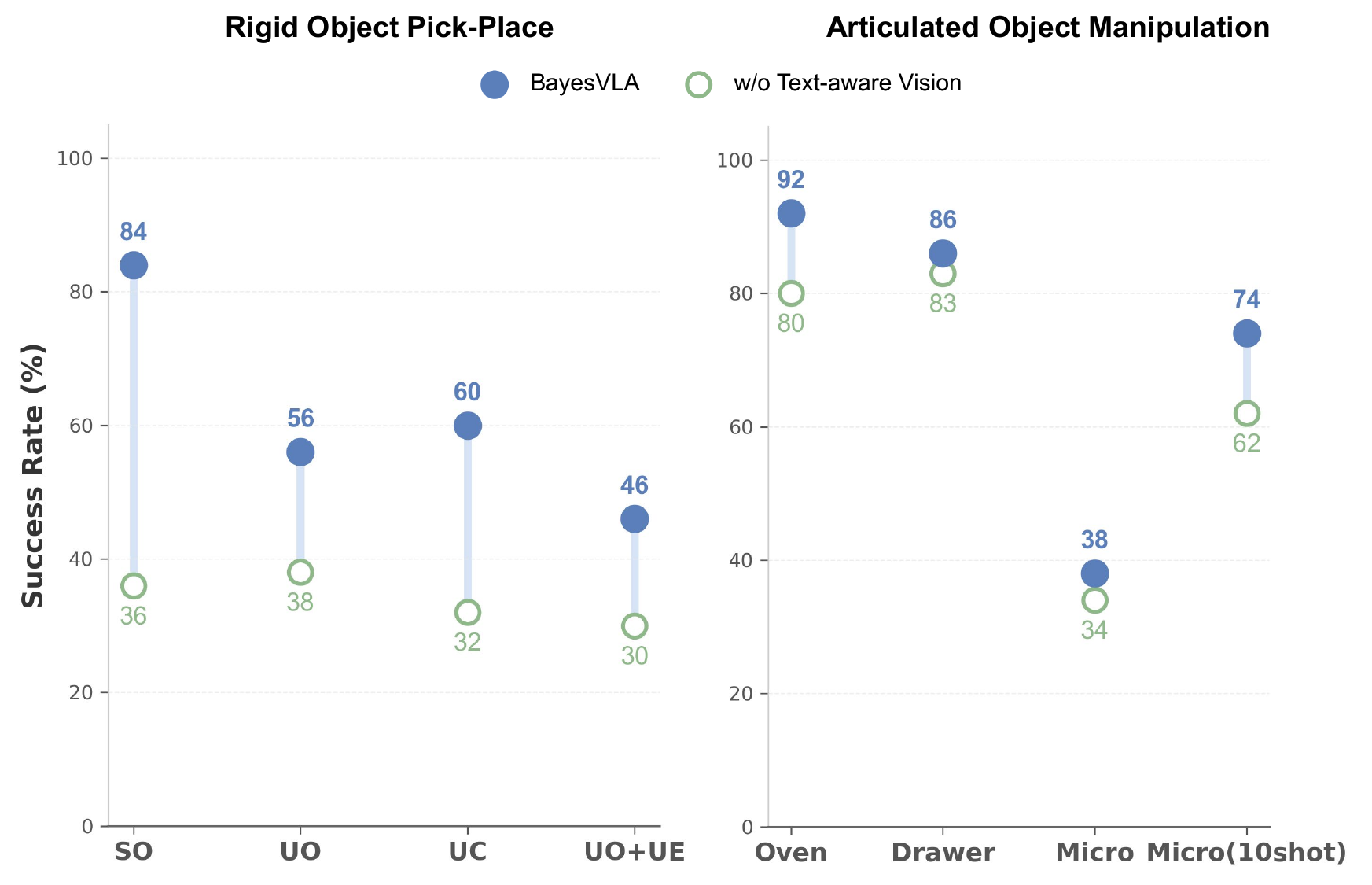}
  \caption{Ablation of text-aware vision tokens in pre-contact likelihood.}
  \label{fig:abl-precontact}
  \vspace{-0.1cm}
\end{figure}

{\bf R1 and R2 Pre-training.} From Table~\ref{tab:libero} to Table~\ref{tab:articulated}, we summarize the effect of different pre-training strategies across all benchmarks. Overall, on the LIBERO benchmark, where the amount of training data is relatively limited, pre-training both the prior and the likelihood~({\it R1}) brings substantial improvements, particularly in object grounding and robustness to environment variations. In contrast, for the rigid object pick-place and articulated object manipulation tasks, where data are more abundant, {\it R2} pre-training provides less pronounced gains in the zero-shot setting. However, it still yields notable improvements in one-shot adaptation scenario~(Table~\ref{tab:1shot-pick-place}). This indicates that our prior pre-training is especially beneficial in few-shot regimes while still preserving zero-shot performance. In addition, these observations collectively suggest that, even without pre-training, sufficient task data allow BayesVLA to learn robust visuomotor priors directly from demonstrations, and maintain zero-shot and few-shot generalization.

{\bf Pre-contact Likelihood: Text-aware Vision.} We remove the text-aware visual feature mechanism from the pre-contact likelihood. As shown in Fig.~\ref{fig:abl-precontact}, this leads to a substantial performance drop, primarily on the rigid object pick-place benchmark, where language grounding across multiple objects is essential. These results show that incorporating text-aware visual features in the pre-contact phase is crucial for grounding task instructions, disambiguating object selection, and resolving visual ambiguities in cluttered scenes. In particular, since the image and text features of CLIP are strictly aligned within a shared feature space, the text-aware visual features become more effective.

\begin{table}[t]
\centering
\caption{Ablation on Post-contact Likelihood.}
\label{tab:abl-postcontact}
\begin{tabular}{lcccc}
\toprule
\multicolumn{5}{c}{\textbf{Rigid Object Pick-Place}} \\
\midrule
Method & SO & UO & UC & UO+UE \\
\midrule
BayesVLA & \textbf{84} & \textbf{56} & \textbf{60} & \textbf{46} \\
Classifier-free Guidance & \textbf{84} & 52 & 48 & \textbf{46}  \\
Classifier Guidance & 0 & 0 & 0 & 0 \\
Token Insertion & 82 & \textbf{56} & 44 & 44 \\
Lang Token Only & \textbf{84} & \textbf{56} & 56 & 40 \\
Qwen Feature & 66 & 48 & 46 & 42 \\

\midrule
\multicolumn{5}{c}{\textbf{Articulated Object Manipulation}} \\
\midrule
\multicolumn{1}{c}{} &
\multicolumn{2}{c}{{SO}} &
\multicolumn{2}{c}{{UO}} \\
\cmidrule(lr){2-3} \cmidrule(lr){4-5}
Method & Oven & Drawer & Micro & Micro~(10shot) \\
\midrule
BayesVLA & \textbf{92} & \textbf{86} & \textbf{38} & \textbf{74} \\
Classifier-free Guidance & 90 & 78 & 33 & 71 \\
Classifier Guidance & 47 & 45 & 30 & 54\\
Token Insertion & 86 & 85 & 26 & \textbf{74} \\
Lang Token Only & 87 & 79 & 23 & 73 \\
Qwen Token & 87 & 81 & 33 & 71 \\
\bottomrule
\end{tabular}
\vspace{-0.1cm}
\end{table}

{\bf Post-contact Likelihood: Latent Adaptation.} We design several variations to validate the advantages of our latent adaptation in post-contact likelihood:
\begin{enumerate}
    \item[A1] {\it Classifier-free Guidance} applies classifier-free guidance for language injection, which directly trains the VLA policy for post-contact without two-stage training.  
    \item[A2] {\it Classifier Guidance} trains an additional classifier to predict the alignment of trajectory and language instruction under visual observation, then guides the diffusion denoising process of VA prior with the classifier gradient. 
    \item[B1] {\it Token Insertion} plugs both text-aware vision tokens and text tokens into the cross-attention layers of VA prior, then tunes these layers as well as the action head.
    \item[C1] {\it Text Token Only} removes text-aware vision tokens and injects merely text tokens during latent adaptation.
    \item[C2] {\it Qwen Token} replaces both vision and text tokens of CLIP with corresponding tokens of Qwen2.5-VL~\cite{bai2025qwen2_5}.
\end{enumerate}

Results are summarized in Table~\ref{tab:abl-postcontact}. Variant A1 and A2 represent different {\it language guidance mechanisms}. Specifically, replacing two-stage training with directly classifier-free guidance leads to notable degradation in the UC setting~($60\%\rightarrow48\%$), which calls for post-contact generalization on objects and language instructions. This further proves the advantage of our two-stage training recipe for prior-likelihood factorization. The classifier-guidance variant exhibits the worst performance, especially on pick-place tasks. In these scenarios, the classifier fails to distingush among multiple objects and is easily distracted by irrelevant cues. This likely stems from the limited generalization. Although the classifier performs well on the training set~(thanks to rollouts from the pre-trained VA prior and label augmentations with aligned and non-aligned instructions), its behavior collapses under even mild distribution shifts. Consequently, the classifier frequently issues incorrect guidance signals, driving the policy off course and eventually causing divergence. The variant B1 showcases another {\it language injection architecture}. For this variant, the semantic-relative tokens should share the same attention layers with the vision tokens of VA prior, indicating worse generalization in unseen container grounding and unseen motion constraint for novel articulated object manipulation. In comparison, introducing additional attention layers to inject both vision and text tokens can complement these capabilities. Finally, variants C1 and C2 compare different {\it text token type}. Injecting only text tokens can achieve comparable performance except for the unseen articulated object manipulation, verifying the benefit of text-ware vision tokens for novel motion generalization. It is also interesting to observe that using Qwen tokens degrades performance across all settings, probably due to the reason that the vision tokens and text tokens are not strictly aligned in feature space. 

\begin{figure*}[t]
  \centering
  \includegraphics[width=\linewidth]{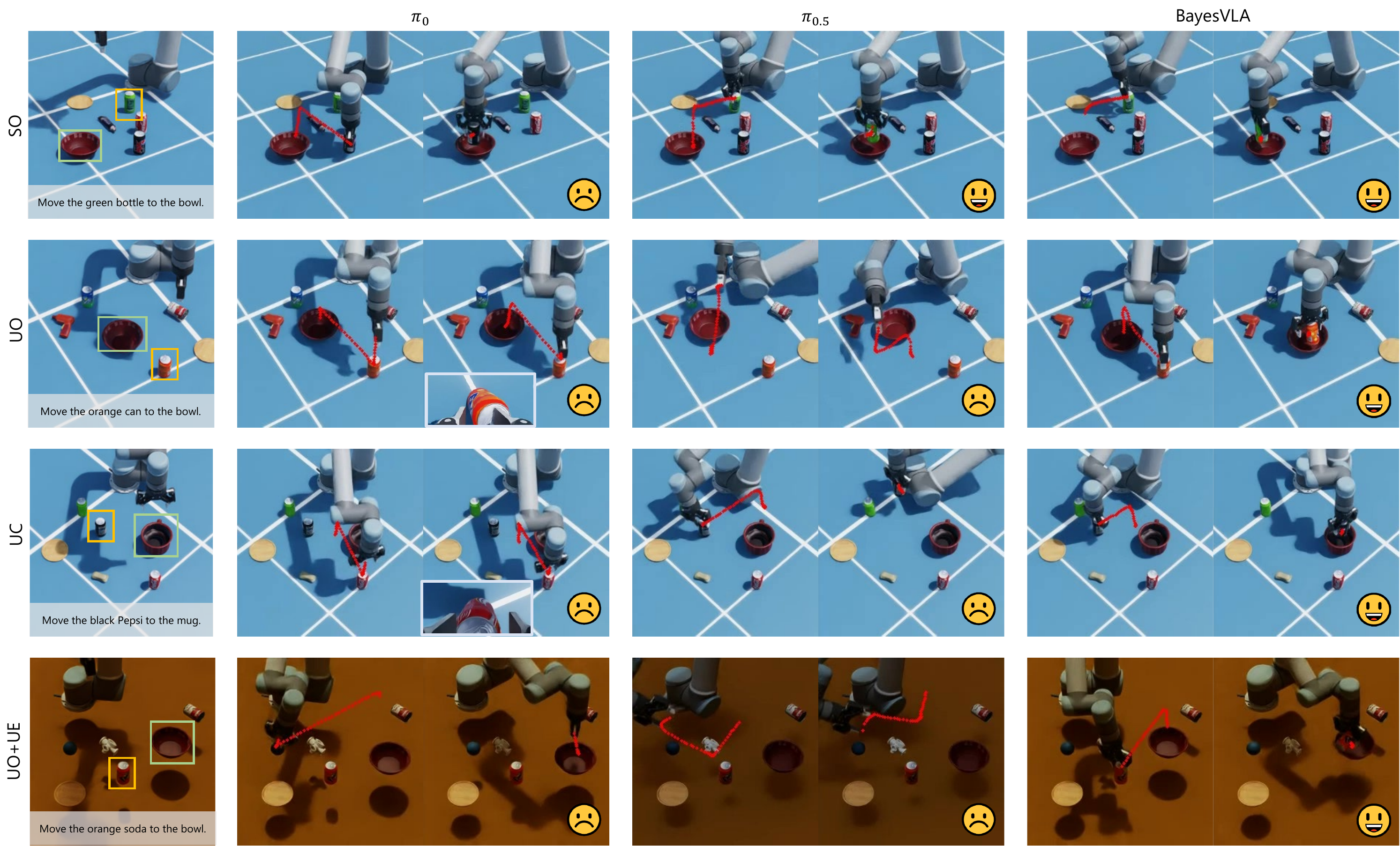}
  \caption{Simulation case studies for rigid object pick-place benchmark. For each case, we show the initial state and language instruction~(left), and the policy performance of $\pi_0$, $\pi_{0.5}$, as well as BayesVLA. We visualize the generated trajectories for each policy with a red point line. }
  \label{fig:sim_case}
  \vspace{-0.1cm}
\end{figure*}

\begin{figure}[t]
  \centering
  \includegraphics[width=\linewidth]{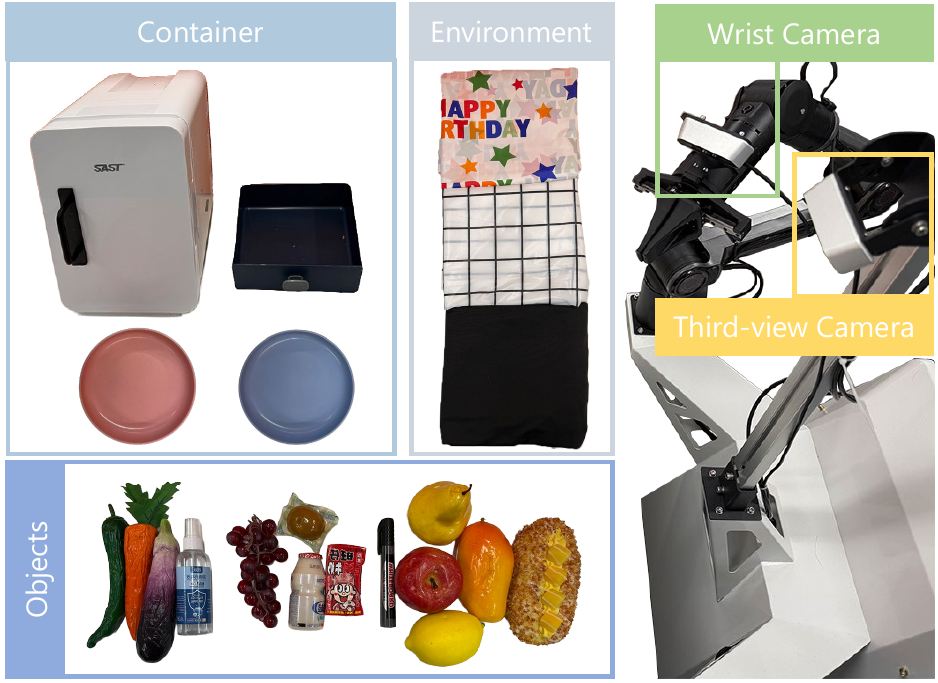}
  \caption{Real-world platform with involved objects and environments.}
  \label{fig:real_platform}
  \vspace{-0.2cm}
\end{figure}

\begin{figure*}[t]
  \centering
  \includegraphics[width=\linewidth]{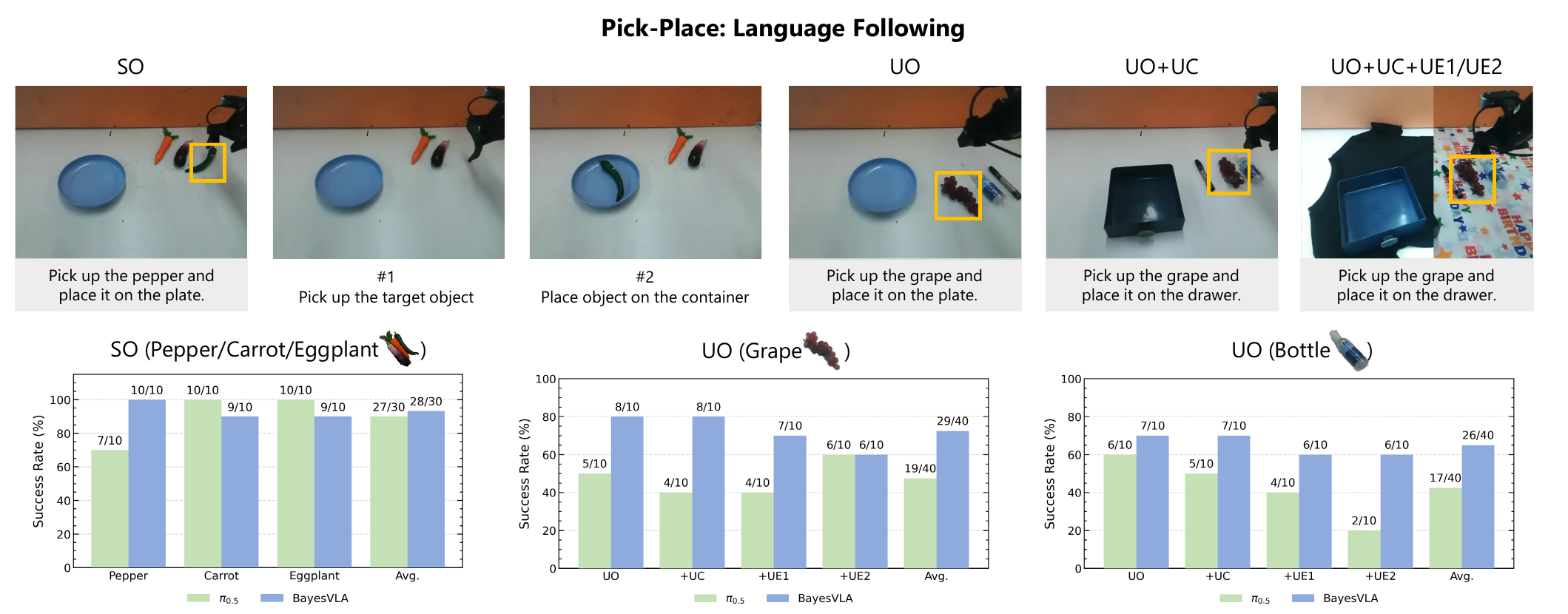}
  \vspace{-0.3cm}
  \caption{Real-world pick-place task description and evaluation results. The representative successful rollout is presented on the top left. For this task, the robot should \#1 pick up the target object, then \#2 place it on the container. We test four settings: seen object with seen container~(SO), unseen object with seen container~(UO), unseen object with unseen container~(UO+UC), and unseen object with unseen container under two distinct unseen backgrounds~(UO+UC+UE1/UE2). Seen objects include pepper, carrot, and eggplant, while unseen objects include grape and bottle.}
  \label{fig:real-pp}
  \vspace{-0.2cm}
\end{figure*}

\begin{figure*}[t]
  \centering
  \includegraphics[width=\linewidth]{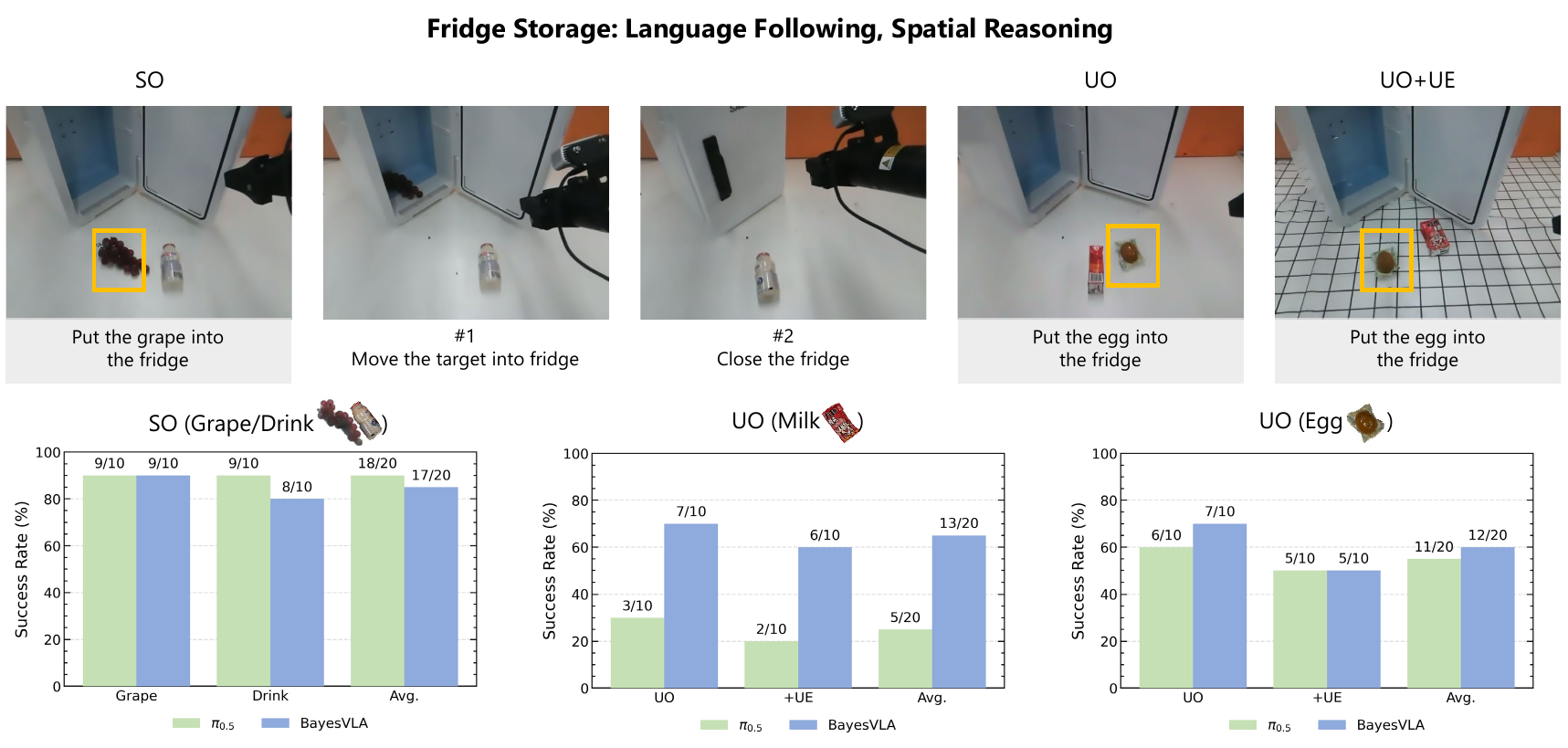}
  \vspace{-0.3cm}
  \caption{Real-world fridge storage task description and valuation results. The representative successful rollout is presented on the top left. For this task, the robot should \#1 move the target object into the fridge, then \#2 close the fridge. We test three settings: seen object with seen container~(SO), unseen object with seen container~(UO), and unseen object with seen container under unseen background~(UO+UE). Seen objects include grape and drink, while unseen objects include mile and egg.}
  \label{fig:real-fridge}
  \vspace{-0.2cm}
\end{figure*}

\begin{figure*}[t]
  \centering
  \includegraphics[width=\linewidth]{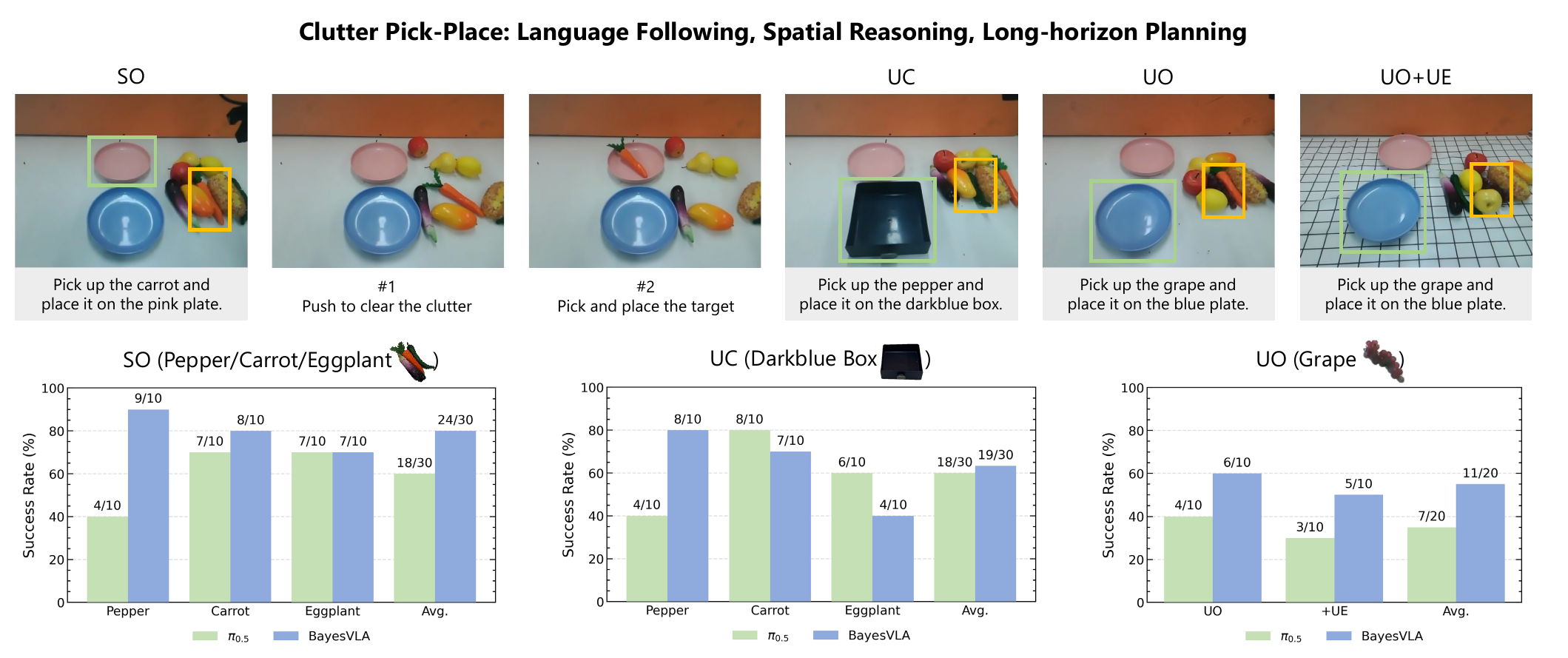}
  \vspace{-0.3cm}
  \caption{Real-world clutter pick-place task description and evaluation results. The representative successful rollout is presented on the top left. For this task, the robot should \#1 push to clear the clutter, then \#2 pick and place the target object. We test four settings: seen object with seen container~(SO), seen object with unseen container~(UC), unseen object with seen container~(UO), and unseen object with seen container under unseen background~(UO+UE). Seen objects include pepper, carrot, and eggplant, while unseen object includes grape.}
  \label{fig:real-clutter-pp}
  \vspace{-0.1cm}
\end{figure*}

\begin{figure*}[t]
  \centering
  \includegraphics[width=\linewidth]{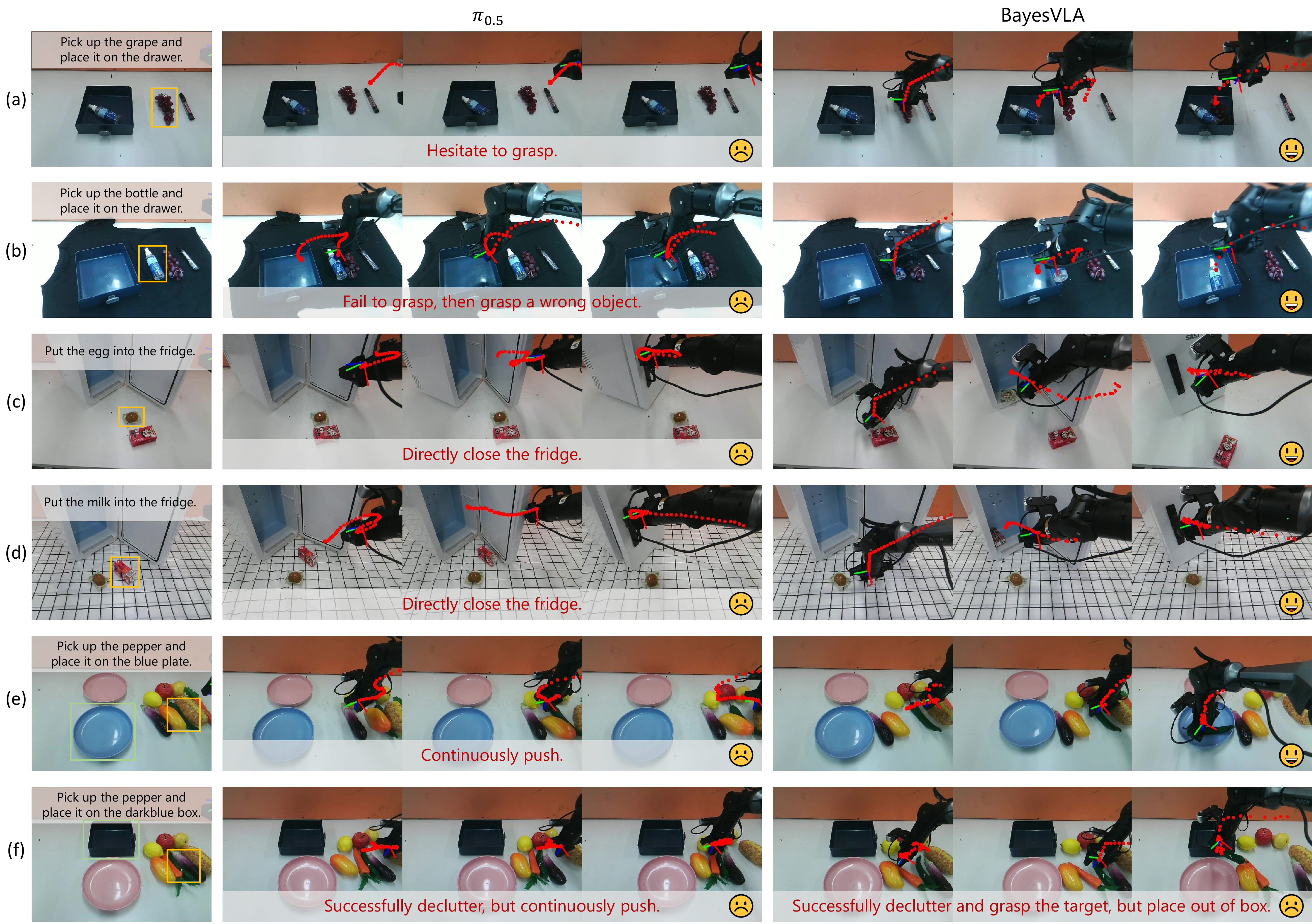}
  \caption{Real-world case studies across all tasks. For each case, we show the initial state and language instruction~(left), and the policy performance of $\pi_{0.5}$ as well as BayesVLA. The generated trajectories for each policy are visualized using red dotted lines, and failure cases are annotated with explanations highlighting the causes of errors.}
  \label{fig:real_case}
  \vspace{-0.1cm}
\end{figure*}

\subsection{Case Studies}

In Fig.~\ref{fig:sim_case}, we visualize representative cases across all settings in the rigid object pick–place benchmark. In the SO case, $\pi_{0}$ grounds the wrong objects, whereas $\pi_{0.5}$ and BayesVLA successfully complete the task with accurate grounding and reliable grasping. In the UO case, although $\pi_{0}$ identifies the correct object, it fails to grasp it due to unintended collision. Besides, $\pi_{0.5}$ is easily distracted by another object with a similar color, repeatedly switching its trajectory between the two. For both UC and UO+UE cases, $\pi_{0}$ again misgrounds the target and collides with objects during grasp attempts. Likewise, $\pi_{0.5}$ struggles to produce trajectories that follow the instruction, gradually deviating from the initially correct direction. Conversely, BayesVLA performs robustly across these diverse variations, indicating the effectiveness of the prior–likelihood factorization and the two-phase design.

\section{Real-world Experiments}

\subsection{Setup}
In this section, we evaluate our system in real-world settings. We use the Agilex Cobot platform, which involves a Piper robot arm and two ORBBEC DaBai cameras for third-view and wrist-view respectively. Both capture RGB-D images with a resolution of 640$\times$480. Fig.~\ref{fig:real_platform} presents our real-world platform as well as the tested objects and backgrounds.

We compare our policy with the baselines $\pi_{0.5}$, which shows better performances than other baselines in our simulation experiments. For each task, we collect 30 tele-operated demonstrations, covering a diversity of language instructions and object layouts. All policies are fine-tuned on each dataset with a batch size of 32. And our policy follows {\it R1} training recipe for all real-world tasks. It is worth noting that the Agilex Cobot configuration is unseen during our pre-training, while $\pi_{0.5}$ was pre-trained in part on the mobile Trossen and mobile ARX platforms, which share similar embodiment structures with Agilex Cobot. In all experiments, $\pi_{0.5}$ is fine-tuned using joint-space actions, consistent with its original pre-training setup on mobile Trossen/ARX. 

\subsection{Pick-place Tasks}
{\bf Task Settings.} For this task, the policy is supposed to pick up the specified object into a container. The language instruction for this task is: ``Pick up the \{object\} and place it on the plate''. Representative successful rollout and example cases of different generalization dimensions are shown in Fig.~\ref{fig:real-pp}. Each policy is evaluated over a total of 110 trials. In the first 30 trials, we evaluate the pick-place performance with the seen objects and container~(SO). For each target object, we conduct 10 trials, where we vary object positions, with other objects serving as distractors. This setup evaluates the language-following capability of policies. The remaining 80 trials evaluate the policy’s generalization performance across multiple dimensions. With two unseen objects as target items, we gradually increase the level of generalization: unseen object with seen container~(UO), unseen object with unseen container~(UO+UC), and unseen object with unseen container under two distinct unseen backgrounds~(UO+UC+UE). Each configuration is tested 10 times. 

{\bf Seen Objects.} As shown in Fig.~\ref{fig:real-pp}, both $\pi_{0.5}$ and BayesVLA achieve strong success rates on the pick-place task with seen objects and container. In particular, our policy consistently completes the task across all target objects, proving both robust language understanding and reliable execution with distractors. 

{\bf Zero-shot Generalization.} The generalization performance in Fig.~\ref{fig:real-pp} highlights the advantages of our approach. Notably, the task complexity gradually increases, from unseen objects with seen container~(UO), to unseen container~(UO+UC), and finally to additional unseen environments~(UO+UC+UE1/UE2). Despite this increasing difficulty, our policy maintains success rates across all levels of generalization. Instead, $\pi_{0.5}$ may fail to pick up the target object, and sometimes mistakenly grasp the non-target object for placement~(Fig.~\ref{fig:real_case}~(b)). Moreover, $\pi_{0.5}$ appears sensitive to environmental disturbances: when the background is replaced with a black cloth, it becomes hesitant during grasping, exhibiting repeated cycles of attempting to pick up the target object and then releasing it. Similarly, we find that $\pi_{0.5}$ is easily disrupted by distractors inside the target container. As shown in Fig.~\ref{fig:real_case}~(a), if a non-target object is initially placed inside the container, $\pi_{0.5}$ often remains completely still, likely misinterpreting the task as already completed. Overall, these results verify that our approach effectively integrates action generation and language-conditioned alignment, preserving the grounding generalization to novel objects and language instructions.

\subsection{Fridge Storage Tasks}
{\bf Task Settings.} For this task, the policy should pick up the specified object into the fridge, and then close the fridge. The language prompt for this task is: ``Put the \{object\} into the fridge''. Representative successful rollout and example cases of different generalization dimensions are shown in Fig.~\ref{fig:real-fridge}. Each policy is evaluated over 60 trials in total. In the first 20 trials, the target objects are seen during training. For the rest 40 trials, we test the generalization of: unseen object~(UO) and unseen object with unseen background~(UO+UE). Each configuration is tested 10 times. 

{\bf Seen Objects.} Both $\pi_{0.5}$ and BayesVLA achieve comparable performance on the fridge storage task, reliably completing both seen-object settings~(Fig.~\ref{fig:real-fridge}). This task requires precise spatial reasoning, as the robot must navigate occlusions imposed by the fridge door. By factorizing the policy into prior and likelihood, even with much less pre-training compared to $\pi_{0.5}$, our policy can capture the effective motion patterns without losing the language following ability.

{\bf Zero-shot Generalization.} Results in Fig.~\ref{fig:real-fridge} show that BayesVLA surpasses $\pi_{0.5}$ across all cases involving unseen objects and backgrounds. A common failure mode for $\pi_{0.5}$ is to directly close the fridge without grasping the target object, as shown in Fig.~\ref{fig:real_case}~(c-d). This means that $\pi_{0.5}$ may fail to correctly identify unseen targets and treat them as distractors. Additionally, under unseen environment settings, $\pi_{0.5}$ sometimes grasps the wrong object, primarily for the unseen object ``milk''. In contrast, BayesVLA presents a more robust performance. Although it occasionally grounds an incorrect object, it reliably completes the tasks. The improved robustness derives from the model’s ability to separate action feasibility~(handled by the prior) from language-conditioned alignment~(guided by the likelihood).

\subsection{Clutter Pick-place Tasks}

{\bf Task Settings.} In this task, the robot cannot directly grasp the object, as it is tightly surrounded by distractors. Therefore, it must first perform push actions to create sufficient clearance for gripper insertion before executing the pick-place actions~\cite{xu2021epg,chen2025clutterdexgrasp}. This requires long-horizon planning across push, pick, and place sub-tasks. Additionally, the task demands identifying the target object among multiple objects and selecting the correct container among several options, emphasizing semantic grounding. The language prompt for this task is: “Pick up the \{object\} and place it on the \{container\},” where the \{object\} denotes the target object to pick, and \{container\} specifies the placement target. Representative successful rollout and example cases of different generalization dimensions are shown in Fig.~\ref{fig:real-clutter-pp}. Each policy is evaluated over 80 trials in total. In the first 30 trials, the target objects and distractors are seen during training. For the rest 50 trials, we test the generalization of: seen object with unseen container~(UC), unseen object with seen container~(UO), and unseen object with unseen background~(UO+UE). Each configuration is tested 10 times. 

{\bf Seen Objects.} We show in Fig.~\ref{fig:real-clutter-pp} that BayesVLA outperforms $\pi_{0.5}$ on seen object trials. $\pi_{0.5}$ sometimes struggles to transfer from the push sub-task to the grasp sub-task, or mistakenly pushes off the target object and picks up the wrong one. We show an example case in Fig.~\ref{fig:real_case}~(e), where $\pi_{0.5}$ does not conduct effective push actions for clutter clearance. The pepper remains largely occupied in the third-view image. Compared to $\pi_{0.5}$, our improved accuracy stems from effective text-aware latent adaptation, which helps focus on the correct object for pushing and grasping. Our method also shows fewer push-grasp transfer hesitations, indicating that the prior effectively encodes general grasping strategies under clutter.

{\bf Zero-shot Generalization.} Fig.~\ref{fig:real-clutter-pp} shows generalization performance under different unseen settings. For seen objects with unseen containers, $\pi_{0.5}$ and BayesVLA achieve comparable performance. $\pi_{0.5}$ shows only a minor performance drop, indicating its robustness to unseen containers. BayesVLA shows a noticeable drop on the eggplant cases, primarily due to its failure to declutter or mistakenly pushing the eggplant out of the workspace. However, when introducing unseen objects, $\pi_{0.5}$ often grounds the wrong object, particularly when the object shares a similar color with the target. For instance, it frequently misidentifies the target ``grape'' as ``eggplant'', both purple but with distinct shapes, resulting in either picking and placing the eggplant or merely pushing around it~(Fig.~\ref{fig:real_case}~(f)). Its performance further degrades under unseen environments: in addition to incorrect grounding, $\pi_{0.5}$ may fail to switch from push to pick or from pick to place. In contrast, BayesVLA makes few wrong-object grounding errors, although it occasionally fails to localize the target container, {\it e.g.}, placing the grape outside the plate. These results confirm that the Bayesian structure provides robustness even in heavily cluttered real-world conditions. Additionally, the model’s latent adaptation allows it to reinterpret visual cues, reducing misidentifications caused by overlapping colors or shapes.

\subsection{Failure Studies}

\begin{figure}[t]
  \centering
  \includegraphics[width=\linewidth]{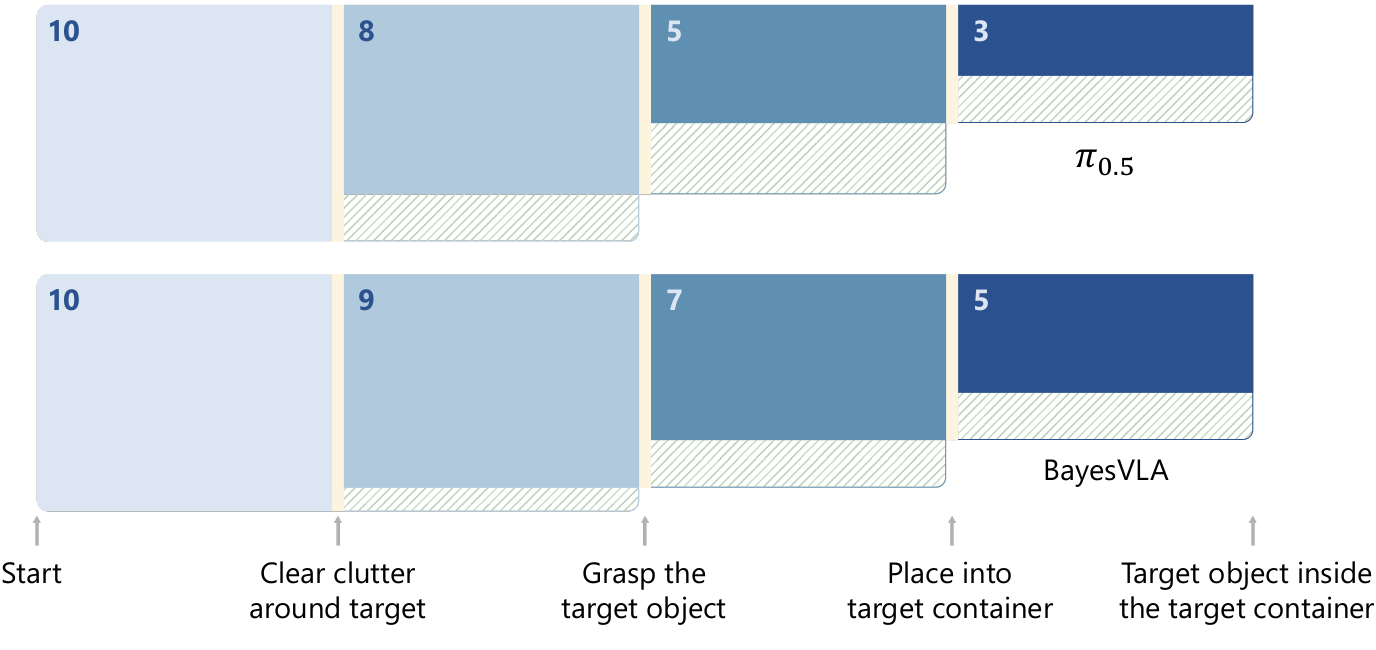}
  \caption{Failure breakdown analysis. We show the Sankey diagram of success~(solid) and failure~(hatch) across the entire rollouts of the UO+UE setting in clutter pick-and-place tasks.}
  \label{fig:failure-analysis}
  \vspace{-0.3cm}
\end{figure}

\begin{figure}[t]
  \centering
  \includegraphics[width=\linewidth]{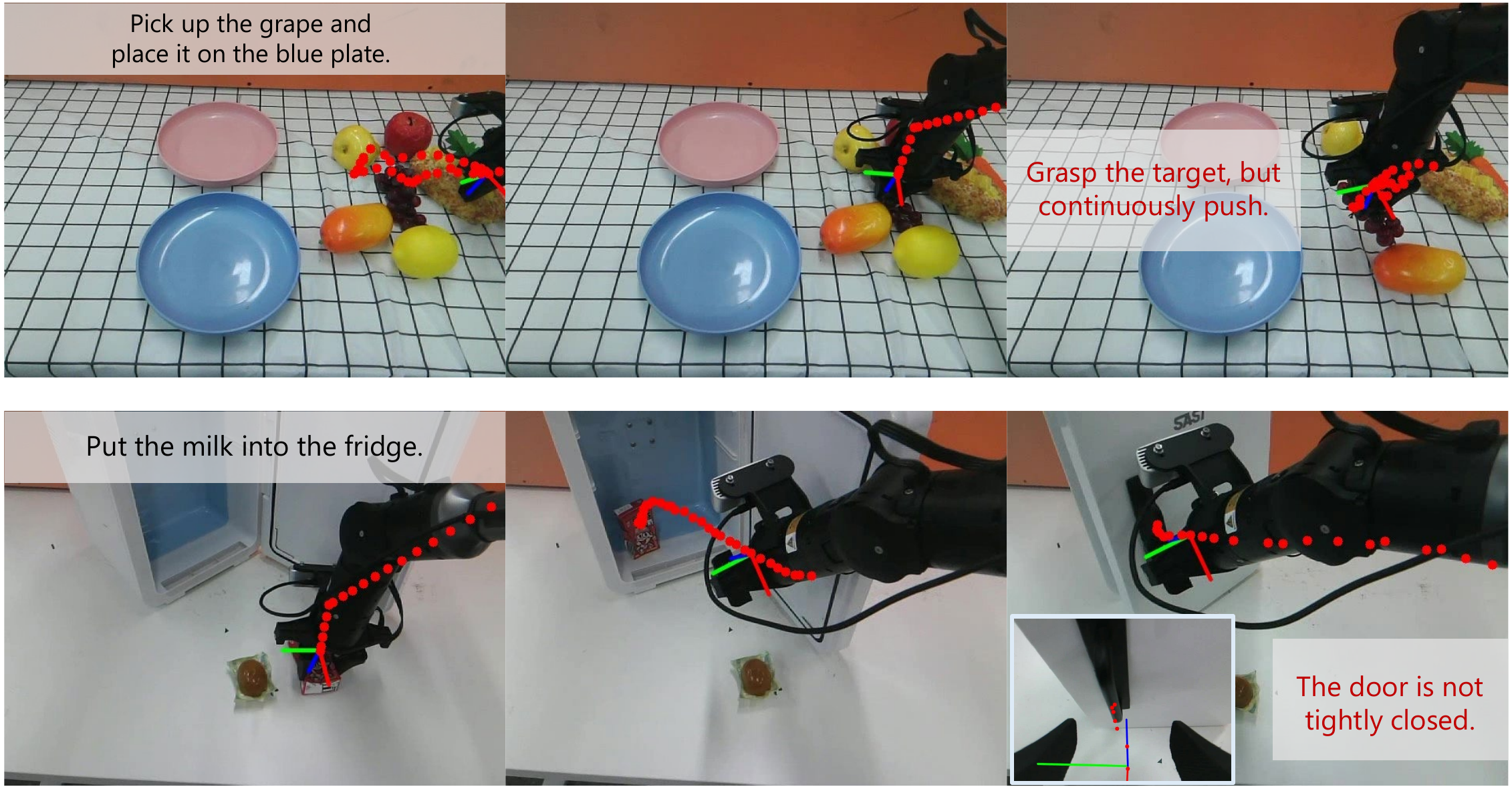}
  \caption{Failure case visualization. For each case, we show the initial state and language instruction, along with the performance of BayesVLA. The generated trajectories for each case are visualized using red dotted lines. Annotated failure reasons highlight the underlying error modes.}
  \label{fig:failure-case}
  \vspace{-0.3cm}
\end{figure}

{\bf Failure Analysis.} In Fig.~\ref{fig:failure-analysis}, we visualize the failure breakdown for the UO+UE setting in clutter pick-and-place tasks. Given an initial cluttered state, both $\pi_{0.5}$ and BayesVLA can robustly clear obstacles around the target. Failures in this sub-task mainly arise when the target and its surrounding objects have highly similar heights, causing the policy to keep pushing without successfully separating them. After decluttering, $\pi_{0.5}$ may hesitate to grasp the target and repeatedly issue push actions, suggesting that it often fails to correctly identify the target object. In addition, $\pi_{0.5}$ twice misclassified an eggplant as a grape, indicating its susceptibility to distractors with similar colors. In contrast, BayesVLA grounds the grape more reliably, though it occasionally fails to execute a successful pick. After the target is finally grasped, $\pi_{0.5}$ may continue pushing while holding the grape, and BayesVLA may exhibit similar behavior. Both policies sometimes fail to switch correctly between different sub-tasks of the long-horizon process.

{\bf Failure Cases.} Fig.~\ref{fig:failure-case} presents concrete failure cases of BayesVLA. The first case corresponds to the UO+UE setting in the clutter pick-place task. BayesVLA successfully declutters the scene and grasps the target object, but fails to transition to the place sub-task. One possible cause is severe occlusion after the object is grasped. The third-view camera is largely blocked by the robot arm, while the wrist camera is obstructed by an unseen object, making it difficult to assess whether the pushing phase should terminate. This limitation is exacerbated by the lack of observation memory. Although the policy conditions on action history, it does not incorporate observation history during action generation. The second case corresponds to the UO setting in the fridge storage task. BayesVLA correctly places the object into the fridge and attempts to close the door, but prematurely considers the task complete before the door is fully shut. Wrist-view visualization shows that the wrist camera provides minimal clues about the door state, forcing the policy to rely primarily on the third-view image. Such reliance becomes problematic in unseen environments, where third-view visual cues may deviate from the training distribution. Incorporating haptic feedback could potentially alleviate this issue by providing more reliable signals of contact for door closure.
\section{Conclusion}

In this work, we propose BayesVLA, a Bayesian modeling framework for Vision-Language-Action policy. By factorizing the policy into a vision-conditioned prior and a language-conditioned likelihood, BayesVLA effectively disentangles multimodal action generation from language grounding. The proposed framework is instantiated for both pre-contact and post-contact actions, where the prior generates contact poses or trajectories, and the likelihood injects language information through alignment. Extensive experiments in simulation and real world validate that BayesVLA achieves superior performance in language following and generalization across diverse tasks. We believe this framework offers a unified and interpretable foundation for advancing scalable, general-purpose robot policies.

{\bf Limitations and Future Work.} In this paper, the Bayesian factorization allows us to leverage pre-trained action foundation models through the pre-contact phase, although it still relies on their quality. Besides, handling long-horizon tasks with complex sequential dependencies remains a major challenge, especially under severe occlusions or ambiguous language instructions. Future work includes unifying the vision–action prior and incorporating large-scale VLMs to improve scalability. Additionally, integrating explicit memory mechanisms and tactile representations could further enhance performance on long-horizon, contact-rich manipulation tasks.

\bibliographystyle{IEEEtran}
\bibliography{IEEEabrv,ref}

\vfill

\end{document}